\pdfoutput=1

\documentclass[11pt]{article}
\usepackage[final]{acl}
\usepackage{colortbl, xcolor} 
\usepackage{times}
\usepackage{latexsym}
\usepackage{amsfonts}
\usepackage{adjustbox}
\usepackage[T1]{fontenc}
\usepackage[utf8]{inputenc}
\usepackage{microtype}
\usepackage{inconsolata}
\usepackage{graphicx}
\usepackage{amsmath}
\usepackage{mathrsfs}
\usepackage{multirow}
\usepackage{multicol}
\usepackage{booktabs}
\usepackage{tcolorbox}
\usepackage[ruled,vlined]{algorithm2e}

\DeclareMathOperator*{\argmax}{arg\,max}

\definecolor{lightgray}{gray}{0.9}
\definecolor{headergray}{gray}{0.8}
\definecolor{darkgray}{gray}{0.6}

\newtcbox{\roundbox}{
    nobeforeafter,    
    box align=base,   
    colback=gray!20,
    colframe=gray!20,
    boxrule=0pt,
    arc=2pt,
    boxsep=0pt,
    left=4pt,
    right=4pt,
    top=2pt,
    bottom=2pt
}

\title{ClaimPKG: Enhancing Claim Verification via Pseudo-Subgraph Generation with Lightweight Specialized LLM}

\author{
  Hoang Pham\thanks{Equal contribution.}, 
  Thanh-Do Nguyen\footnotemark[1], 
  Khac-Hoai Nam Bui\thanks{Corresponding author.} \\
  Viettel Artificial Intelligence and Data Services Center, \\
  Viettel Group, Vietnam \\
  \{hoangpv4, dont15, nambkh\}@viettel.com.vn
}

\begin{document}
\maketitle
\begin{abstract}
Integrating knowledge graphs (KGs) to enhance the reasoning capabilities of large language models (LLMs) is an emerging research challenge in claim verification. While KGs provide structured, semantically rich representations well-suited for reasoning, most existing verification methods rely on unstructured text corpora, limiting their ability to effectively leverage KGs. Additionally, despite possessing strong reasoning abilities, modern LLMs struggle with multi-step modular pipelines and reasoning over KGs without adaptation. To address these challenges, we propose ClaimPKG\footnote{https://github.com/HoangHoang1408/ClaimPKG}, an end-to-end framework that seamlessly integrates LLM reasoning with structured knowledge from KGs. Specifically, the main idea of ClaimPKG is to employ a lightweight, specialized LLM to represent the input claim as pseudo-subgraphs, guiding a dedicated subgraph retrieval module to identify relevant KG subgraphs. These retrieved subgraphs are then processed by a general-purpose LLM to produce the final verdict and justification. Extensive experiments on the FactKG dataset demonstrate that ClaimPKG achieves state-of-the-art performance, outperforming strong baselines in this research field by 9\%-12\% accuracy points across multiple categories. Furthermore, ClaimPKG exhibits zero-shot generalizability to unstructured datasets such as HoVer and FEVEROUS, effectively combining structured knowledge from KGs with LLM reasoning across various LLM backbones.
\end{abstract}
\section{Introduction}
\begin{figure}[!h]
  \centering
  \includegraphics[width=0.8\columnwidth]{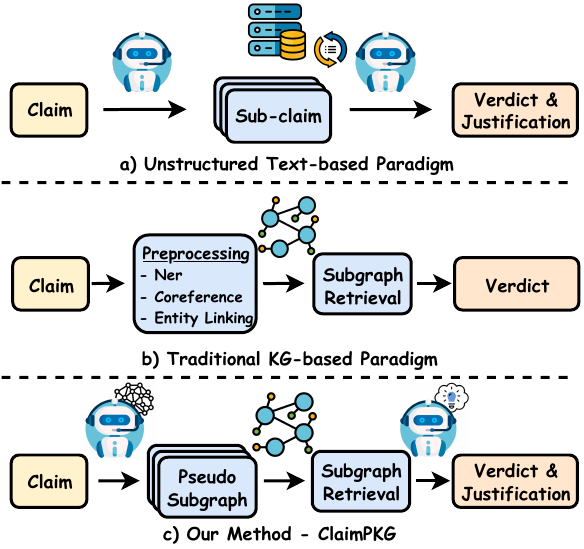}
  \caption{Different claim verification paradigms: (a) Unstructured Text-based methods focusing on claim decomposition and sequential reasoning over text, (b) KG-based methods facing challenges in entity resolution and structured reasoning, and (c) ClaimPKG's unified framework with specialized modules for pseudo-subgraph generation, retrieval, and general reasoning.}
  \label{fig:comparison}
\end{figure}
In today's rapidly evolving information landscape, distinguishing fact from misinformation is becoming more challenging, especially with the rise of AI-generated content. Robust claim verification systems, leveraging NLP methods to automatically assess the veracity of claims \cite{task_definition_2,task_definition_3,task_definition_1}, are essential to ensure information reliability. Effective methods require not only accuracy but also transparency, necessitating strong reasoning to identify evidence and provide clear justifications \cite{programfc}.


Most existing verification approaches focus on unstructured text corpora, using techniques like chain-of-thought (CoT) reasoning \cite{chainofthought} to break down claims for verification. Approaches like ProgramFC \cite{programfc} and FOLK \cite{folk} employ modular pipelines to verify claims against text-based knowledge bases (Figure \ref{fig:comparison}(a)). However, the inherent limitations of text representation pose challenges. Specifically, ambiguous entity references and complex multi-hop relationships make it difficult to perform rigorous verification against unstructured text.

In contrast, Knowledge Graphs (KGs) provide structured relationships for effective reasoning \cite{reasoningongraph, thinkongraph}, yet their use in claim verification remains limited. Existing KG-based approaches (Figure \ref{fig:comparison}(b)) \cite{factkg, gear, kg_gpt} lack end-to-end solutions, often requiring pre-extracted entities via modules like entity or relation extraction. Meanwhile, despite excelling at general reasoning, LLMs struggle with KG-specific tasks like entity resolution and multi-hop reasoning \cite{autoregressiveentityretrieval, feverous}, suggesting the need for a system combining LLM capabilities with KG-based inference.

Overall, solving claim verification problems is hindered by following major limitations: \textit{(1) Entity Ambiguity:} Systems must accurately disambiguate entities within claims to identify relevant evidence \cite{feverous}; \textit{(2) Multihop Reasoning:} Complex claims often require reasoning across multiple evidence from different sources \cite{programfc, folk}; and \textit{(3) Limited integration of KGs and LLMs:} Current approaches are underexploring the potential of combining the application of structured representation with strong inference capabilities of LLMs \cite{kg_gpt}.

To address these challenges, we propose ClaimPKG (Claim Verification using Pseudo-Subgraph in Knowledge Graphs), a novel end-to-end framework that synergizes the adaptability and generalization strengths of LLMs with the structured and rigorous representation of KGs to enable robust and transparent claim verification. As specified in Figure \ref{fig:comparison}(c), ClaimPKG operates through three phases: (1) \textbf{Pseudo-Subgraphs Generation}: A KG-specialized lightweight LLM generates pseudo subgraphs as the representations of input claims under a Trie-based KG-Entity Constraint, ensuring the correctness of extracted entities; (2) \textbf{Subgraphs Retrieval}: A retrieval algorithm considers generated pseudo subgraphs as queries to identify actual relevant KG subgraphs as evidence;
and (3) \textbf{General Reasoning}: A general-purpose LLM reasons over the retrieved KG subgraphs to produce the verdict and human-readable justifications. Through extensive experiments on the FactKG dataset, ClaimPKG achieves state-of-the-art performance, demonstrating its effectiveness over various claim types with a small number of training samples. Furthermore, its zero-shot generalizability to unstructured datasets (HoVer, FEVEROUS) highlights its robustness.

Our contributions can be summarized as follows: 
(1) We introduce ClaimPKG, a holistic framework that integrates LLMs and KGs for accurate and interpretable claim verification, handling various types of claims in a unified manner; 
(2) We develop a lightweight specialized LLM with its according decoding algorithm for pseudo-subgraph generation and pair it with general-purpose LLMs to achieve robust reasoning; and 
(3) We validate the effectiveness of ClaimPKG through extensive experiments, achieving state-of-the-art performance on structure-based datasets and generalizing to unstructure-based datasets.
\section{Related Work}
\textbf{Claim Verification Approaches.} Claim verification systems utilize knowledge bases that can be categorized into unstructured and structured formats. In the unstructured domain, text-based verification methods predominate, with systems designed to verify claims against textual evidence, as demonstrated in the FEVER dataset \cite{fever}. Recent advances have focused on handling specialized verification scenarios, including ambiguous question-answer pairs \cite{faviq}, detecting factual changes \cite{vitamin-c}, and processing multiple documents concurrently \cite{hover}. For structured verification, research has primarily focused on tables and graphs, with early work developing specialized architectures: graph neural networks for knowledge graph processing \cite{graph-review}, table-specific transformers \cite{tapas}, and tree-structured decoders for hierarchical data \cite{rat-sql}.
\\[0.1cm]
\textbf{Claim Verification over Knowledge Graphs (KGs).} The emergence of Large Language Models (LLMs) has simplified direct reasoning over textual corpora for claim verification, as demonstrated by ProgramFC \cite{programfc} and FOLK \cite{folk}. However, structured data sources like tables and graphs can provide more grounded and robust verification results \cite{factkg}. Knowledge graphs are particularly advantageous as they enable explicit representation of reasoning processes through logical rules over nodes and edges. FactKG \cite{factkg} established a foundation in this direction by introducing a comprehensive dataset for evaluating modern verification methods. KG-GPT \cite{kg_gpt} followed this work by demonstrating performance gains through a pipeline that performs sentence decomposition, subgraph retrieval, and logical inference. Additionally, while not directly addressing claim verification, StructGPT \cite{struct-gpt} and RoG \cite{reasoningongraph} achieved promising results in related tasks (e.g., Knowledge Base Question Answering) by collecting relevant evidence, such as subgraphs in KGs, then leveraging LLMs for complex reasoning in particular scenarios. 

\section{Preliminary}
\textbf{Knowledge Graph:} Knowledge Graph (KG) \( \mathcal{G} \) represents facts as triplets of format \( t = (e, r, e') \), where entities \( e, e' \in \mathcal{E} \) are connected by a relation \( r \in \mathcal{R} \); $r$ can also be referred as $r(e, e')$.
\\[0.1cm]
\textbf{Claim Verification:} Given a claim \( c \), a verification model \( \mathcal{F} \) determines its veracity as \textit{Supported} or \textit{Refuted} based on an external knowledge base \( \mathcal{K} \), while also providing a justification \( j \) to explain the predicted label. This work specifically considers the scenario where \( \mathcal{K} \) is structured as a Knowledge Graph \( \mathcal{G} \), enabling reasoning over graph knowledge to infer \( v \) and \( j \). Formally, the verification process is defined as: \((v, j) = \mathcal{F}(c, \mathcal{G}).\)
\\[0.1cm]
\textbf{Trie-based Constrained Decoding:} A Trie \cite{trie} indexes predefined token sequences, where each root-to-node path represents a prefix. During LLM generation, this structure restricts token selection to only valid Trie paths, ensuring reliable output.
\section{ClaimPKG}
\subsection{Formulation of ClaimPKG}
\label{claimpkg_formulation}
\begin{figure*}[!h]
  \centering
  \includegraphics[width=0.99\textwidth]{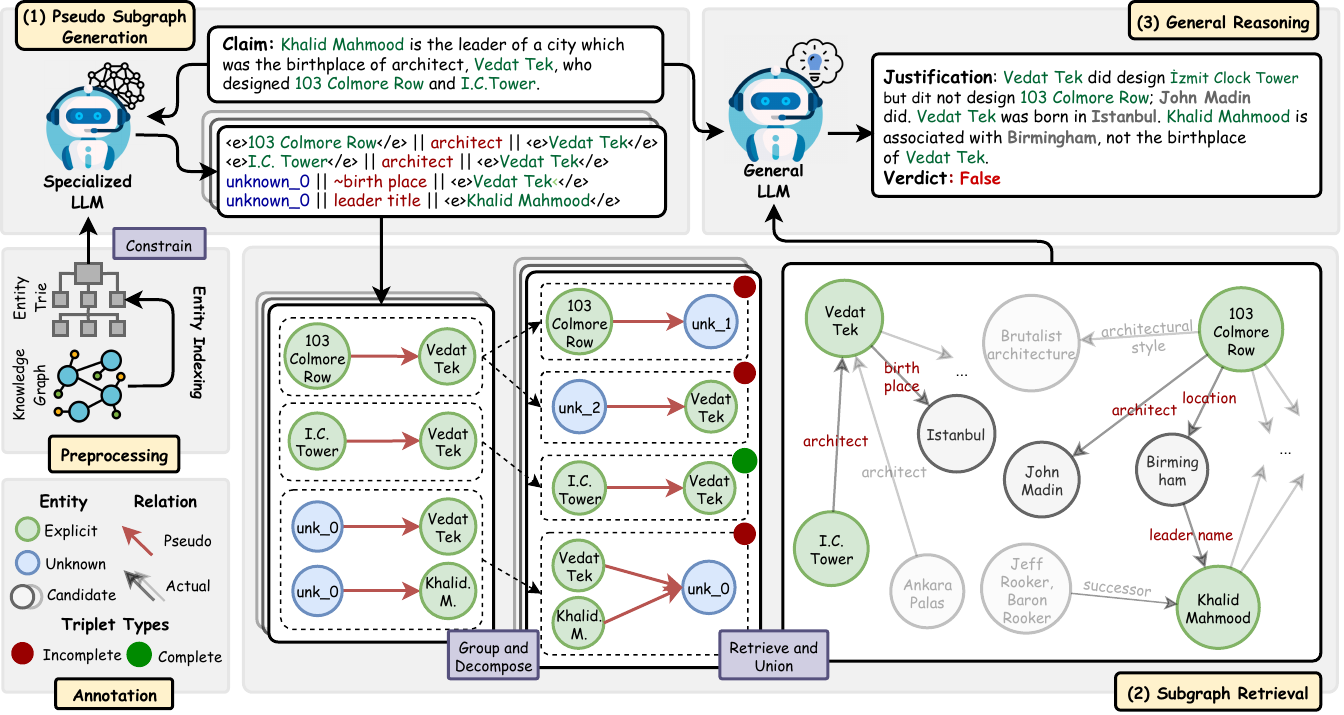}
  \caption{Illustration of the ClaimPKG for claim verification. The framework consists of three key modules: (1) Pseudo-subgraph Generation, constructing representative subgraphs; (2) Subgraph Retrieval, selecting the most pertinent KG subgraphs; and (3) General Reasoning, integrating them for accurate and interpretable verification.}
  \label{fig:main_pipeline}
\end{figure*}
We formulate the ClaimPKG framework using a probabilistic approach. Given a claim $c$ and a prebuilt KG $\mathcal{G}$, our objective is to model the distribution $p_\theta(v,j|c,\mathcal{G})$, where $v$ denotes the verdict and $j$ the justification. However, direct computation for this distribution is infeasible as reasoning over the entire KG is not practical given its large size. To address this, we propose to select $\mathcal{S}_c$, a subgraph of $\mathcal{G}$ relevant to $c$ containing necessary information to derive our target distribution. Treating $\mathcal{S}_c$ as a latent variable, $p_\theta(v,j|c,\mathcal{G})$ is decomposed as:
\begin{equation}
    \textstyle p_\theta(v,j|c,\mathcal{G}) = \sum\limits_{\mathcal{S}_c} p_\theta(v,j|c,\mathcal{S}_c) p_\theta(\mathcal{S}_c|c,\mathcal{G})
\label{eq:1st_decompose}
\end{equation}
where $p_\theta(\mathcal{S}_c|c,\mathcal{G})$ models the subgraph selection, and $p_\theta(v,j|c,\mathcal{S}_c)$ models the generator of the verdict and justification given $\mathcal{S}_c$. However, direct computation of $p_\theta(\mathcal{S}_c|c,\mathcal{G})$ is challenging due to modality mismatch between the input $c$ (text) and the target $\mathcal{S}_c$ (graph structure), hindering the employment of retrieval methods for $\mathcal{S}_c$.
To bridge this gap, we decompose the subgraph selection into:
\begin{equation}
    \textstyle p_\theta(\mathcal{S}_c|c,\mathcal{G}) =  \sum\limits_{\mathcal{P}_c} p_\theta(\mathcal{S}_c|\mathcal{P}_c,\mathcal{G}) p_\theta(\mathcal{P}_c|c,\mathcal{G})
\label{eq:2nd_decompose}
\end{equation}
where $p_\theta(\mathcal{P}_c|c,\mathcal{G})$ models the generation of the graph representation $\mathcal{P}_c$, which we refer as ``pseudo subgraph'', from a textual claim $c$, and $p_\theta(\mathcal{S}_c|\mathcal{P}_c,\mathcal{G})$ models the distribution over relevant subgraphs $\mathcal{S}_c$ given $\mathcal{P}_c$. While equations \ref{eq:1st_decompose} and \ref{eq:2nd_decompose} establish our theoretical framework for ClaimPKG, computing exact probabilities by summing over all possible $(\mathcal{S}_c, \mathcal{P}_c)$ pairs is intractable. Addressing this we propose two approximations: (1) We infer the veracity using only the most relevant subgraph $\mathcal{S}^*_c$:
\begin{equation}
    \textstyle (v^*,j^*) \sim p_\theta(v,j|c,\mathcal{S}^*_c)
\label{eq:1st_approx}
\end{equation}
(2) We assume each generated pseudo-subgraph is reasonable with a high probability, allowing us to approximate the subgraph selection in \ref{eq:2nd_decompose} as:
\begin{equation}
    \textstyle \mathcal{S}^{(i)}_c = \arg\max p_\theta(\mathcal{S}_c|\mathcal{P}^{(i)}_c,\mathcal{G})
\label{eq:2nd_approx}
\end{equation}
with $\mathcal{P}^{(i)}_c$ is the $ith$ pseudo-graph generation. We then construct $\mathcal{S}^*_c$ by aggregating multiple sampled subgraphs, specifically $\mathcal{S}^{*}_c = \bigcup_{} \mathcal{S}^{(i)}_c$.

These approximations lead ClaimPKG to comprise 3 key modules as depicted in Figure {\ref{fig:main_pipeline}}: (1) \textit{Pseudo Subgraph Generation} to generate graph representations $\mathcal{P}_c$'s given claim $c$; (2) \textit{Subgraph Retrieval} to retrieve relevant evidence subgraph $\mathcal{S}^*_c$; and (3) \textit{General Reasoning} to generate final verdict $v$ and justification $j$. The inference procedure is described as follows:

\begin{tcolorbox}[
    colback=gray!10,
    colframe=gray!20,
    boxrule=0.4pt,
    title=Inference Procedure of ClaimPKG,
    fonttitle=\bfseries,
    coltitle=black,  
    left=4pt,     
    right=4 pt,    
    top=4pt,       
    bottom=4pt,     
    label=fig:inference,
    label type=figure,
]
\textbf{Preprocessing:} Index the KG $\mathcal{G}$ into an Entity Trie for effective entity lookup.
\\
\textbf{1. Pseudo Subgraph Generation:} Generate multiple graph representations (pseudo subgraphs) $\mathbb{P}_c=\{\mathcal{P}^{(i)}_c\}^N_{i=1}$ from claim $c$, using a specialized LLM with beam search and Entity-Trie constraints.
\\
\textbf{2. Subgraph Retrieval:} Use each pseudo graph in 
$\mathbb{P}_c$
for querying the most respective relevant subgraph $\mathcal{S}^{(i)}_c$ in the KG $\mathcal{G}$, resulting in a set of $\{\mathcal{S}^{(i)}_c\}^N_{i=1}$ following Equation \ref{eq:2nd_approx}, then aggregate them to form $\mathcal{S}^*_c = \bigcup_{i=1}^N \mathcal{S}^{(i)}_c$.
\\
\textbf{3. General Reasoning:} Employ a general-purpose LLM to reason veracity $(v^*, j^*) \sim p_\theta(v,j|c,\mathcal{S}^*_c)$ following Equation \ref{eq:1st_approx}.
\end{tcolorbox}
The subsequent sections provide details about each component in the ClaimPKG framework.



\subsection{Pseudo Subgraph Generation}
The first step to effectively verify a claim is to understand its content thoroughly and represent it in a format compatible with the KG.
Since evidence comes from KG, representing claims in the graph format is crucial, which captures hypothetical relations among entities in an effective way that enables effective comparisons with KG subgraphs for evidence retrieval.  
However, this process faces \textbf{two main challenges:} (1) handling ambiguity resolution and multi-hop reasoning, and (2) ensuring accurate entity extraction from the claim.  
\\[0.1cm]
\textbf{Specialized LLM.} To address the first challenge, the Pseudo Subgraph Generation module employs a lightweight model optimized for processing input claims. Following \cite{joint_extract_1, joint_extract_2}, the model is trained to jointly extract entities and their corresponding relations from a claim \( c \). Specifically, from $c$ the model constructs a pseudo subgraph $\mathcal{P}_c$ comprising triplets in the form of \roundbox{$\text{head\_entity} || \text{relation} || \text{tail\_entity}$} (illustrated in Figure \ref{fig:main_pipeline}). To ensure the generated subgraph can identify entities requiring ambiguity resolution and multi-hop reasoning, we employ a specialized \textbf{annotation mechanism}: when the claim references an entity indirectly—either without explicit naming or through relations to other entities—we denote it as \roundbox{$\text{unknown}\_i$}, with the index $i$ to keep track of different entities. 
This notation effectively 
signals the need for further disambiguation and reasoning within the KG in subsequent steps. Training details enabling this annotation strategy are presented in Appendix \ref{training_data_annotation}.
\\[0.12cm]
\textbf{Trie-Constrained Decoding.} For the second challenge, we develop a constrained decoding algorithm with an Entity Trie inspired by \cite{autoregressiveentityretrieval}. We construct a trie $\mathcal{T}$ from the KG's entity set \( \mathcal{E} = \{e_1, e_2, ...\} \). The specialized LLM generates entities using special tokens \( \langle e \rangle \) and \( \langle /e \rangle \) to mark entity boundaries. When \( \langle e \rangle \) is generated, the decoding process restricts token selection based on $\mathcal{T}$ until \( \langle /e \rangle \) is produced, ensuring all generated entities exist in the KG. Outside such boundaries, the model generates relations by sampling from an unconstrained original token distribution. This mechanism ensures entity reliability while preserving flexible relation extraction \cite{graphrag}.
\\[0.12cm]
\textbf{Multiple Representations.} In order to capture different semantic views of a claim, we employ beam search along with the described sampling strategy, which is proved to improve the coverage of extracted triplets (table \ref{tab:beam_metrics}), resulting in multiple representations $ \mathbb{P}_c = \{\mathcal{P}^{(i)}_c\}^N_{i=1}$ for an input claim.

In summary, each of the claim's graph representations satisfies following properties: (1) effectively capture the underlying graph structure of that claim, and (2) correctly align with the KG's entities.  

\subsection{Subgraph Retrieval}
\label{section:subgraph_retrieval}
The second component of ClaimPKG involves retrieving relevant KG subgraphs as evidence by using a dedicated algorithm that matches the pseudo-subgraphs $\mathcal{P}_c$'s from the previous step to actual subgraphs in the KG. We present the high-level description of our algorithm here, while its complete formulation is detailed in Appendix \ref{appendix:subgraph_retrieval_algorithm_details}. We categorize triplets in a $\mathcal{P}_c$ into: (1) \textit{Incomplete} triplets, where either the head or tail entity is marked as unknown, and (2) \textit{Complete} triplets, where both head and tail entities are explicitly identified.
\\[0.1cm]
\textbf{Relation Scoring Function:} We define a function $\text{Sim}(r_1, r_2)$ to quantify the similarity between two relations, where a higher score indicates greater similarity. This function can be instantiated via various mechanisms (e.g., embedding similarity, re-ranking, fuzzy matching, etc.).
\\[0.1cm]
\textbf{Incomplete Triplets Retrieval:} Our goal is to identify evidence (actual triplets in the KG) to inform us about entities marked as unknown and their respective relations with explicit entities in the pseudo-subgraphs. First, for a $\mathcal{P}_c$, we group triplets sharing the same unknown entity $u$ into a group $g$ (e.g., in Figure~\ref{fig:main_pipeline}, triplets associated with \roundbox{unknown\_$0$} are grouped together). Subsequently, for each group $g$ characterized by the unknown entity $u$, we denote: $\mathcal{E}_u = \{e_{u1}, \dots, e_{un}\}$ as entities directly connected to $u$ in the pseudo-subgraph $\mathcal{P}_c$ and $\mathcal{R}_u = \{r_{u1}, \dots, r_{un}\}$ as relations from $u$ to corresponding entities in $\mathcal{E}_c$. In $g$, for each explicit entity $e_{ui} \in \mathcal{E}_u$, we first retrieve candidate set $C_{ui}=\{e^c_{i1},\dots,e^c_{im}\}$ containing all entities connected to $e_{ui}$ in the KG, then collect all candidate sets into $\mathcal{C}_u = \{C_{u1},\dots,C_{un}\}$. 

To determine the best candidates for resolving $u$, we propose an Entity Scoring mechanism, which is based on \textbf{two assumptions}: (1) since $u$ has pseudo relations with all entities in $\mathcal{E}_u$, a candidate $e^c$ connected to more entities in $\mathcal{E}_u$ is more likely to resolve $u$; and (2) because every information related to $e_{ui}$ and $u$ is crucial to verify the initial claim, each candidate set $C_{ui}$ must contribute to the final verification.
Note that an entity can appear in multiple candidate sets, hence we compute a ``global'' score for each $e^c_{ij}$ in a candidate set $C_{ui}$:
\begin{equation}
     \textstyle score(e^c_{ij}) = \sum_{r}^{R^u_{ij}}{\text{Sim}(r_{ui}, r)}
\label{eq:scoring}
\end{equation}
with $R^u_{ij} = \bigcup^{|\mathcal{E}_u|}_{i=1}\{r(e_{ui},e^c_{ij}) \mid \text{if } e^c_{ij} \in C_{ui}\}$, the set of all relations across candidate sets appearing in $\mathcal{C}_u$ that connect $e^c_{ij}$ with an $e_{ui}$. Subsequently, to construct the set $T_u$ of most relevant triplets to a group $g$, we employ a ranking function as follows:
\begin{equation}
    T_{u} = \bigcup_{i=1}^{|\mathcal{C}_u|}\argmax_{triplet,k_1}\{\pi_{ij} \mid j \leq \lvert C_{ui} \rvert\} 
\label{eq:ranking}
\end{equation}
with $\pi_{ij}$ is simply $score(e^c_{ij})$ and $(triplet,k_1)$ denotes the selection of top $k_1$ triplets $(e_{ui},r,e^c)$ having the highest global scores from each set in $\mathcal{C}_u$.

While equation \ref{eq:scoring} ensures candidates appearing in multiple candidate sets and having high similar scores are prioritized, equation \ref{eq:ranking} ensures every entity in $\mathcal{E}_u$ has at least $k_1$ triplets, both of which make use of assumptions (1) and (2).
\\[0.1cm]
\textbf{Complete Triplets Retrieval:}
For each triplet $(e_1, r, e_2)$ in a $\mathcal{P}_c$, we first find top $k_2$ similar relations between $e_1$ and $e_2$ in the KG $\mathcal{G}$ using the Sim function. If no direct connection exists (e.g., "103 Colmore Row" and "Vedat Tek" as shown in figure \ref{fig:main_pipeline}), the triplet is decomposed into two: $(e_1, r, \text{unknown}_0)$ and $(\text{unknown}_0, r, e_2)$. These are then handled via Incomplete Triplets Retrieval.
\\[0.1cm]
\textbf{Subgraph Union:} In summary, for an input claim $c$, multiple pseudo-graphs are generated, containing \textit{complete} and \textit{incomplete} triplets. These triplets undergo processing to handle shared unknown entities and identified entities that are not connected in the KG $\mathcal{G}$, and are used to query $\mathcal{G}$ for relevant triplets.  All retrieved evidence triplets are aggregated into a final subgraph $\mathcal{S}^*_c$, serving as the evidence for the final component of ClaimPKG.

\subsection{General Reasoning}
The \textit{General Reasoning} module concludes the ClaimPKG framework by determining claim veracity through reasoning over input claim $c$ and retrieved evidence subgraph $\mathcal{S}^*_c$. As complex tasks, especially claim verification, require deliberate chain-of-thought reasoning \cite{hover, self_consistency}, we use a general-purpose LLM to analyze \(c\) and \(\mathcal{S}^*_c\). Using carefully designed prompts (Figure \ref{fig:prompt_general_reasoning}), the module generates a natural language justification \(j\) and verdict \(v\). Expanded from equation \ref{eq:1st_approx}, this step is formalized as:
\begin{equation}
    \textstyle p_{\theta}(v,j | c, \mathcal{S}^*_c) = p_{\theta}(v | c, j, \mathcal{S}^*_c) p_{\theta}(j | c, \mathcal{S}^*_c)
\end{equation}
where \(p(j | c, \mathcal{S}^*_c)\) produces the justification and \(p(v | c, j, \mathcal{S}^*_c)\) determines veracity. This model-agnostic design enables integration with state-of-the-art LLMs (e.g., Llama, Qwen and GPT4) for zero-shot reasoning.
\section{Experiments}
\subsection{Experimental Setup}
\textbf{Datasets.} Our primary benchmark is the FactKG dataset \cite{factkg}, designed for claim verification over the DBpedia KG \cite{dbpedia}. It consists of 108K claims grounded in DBpedia and labelled as either \textit{SUPPORTED} or \textit{REFUTED}. The claims span five distinct categories: One-hop, Conjunction, Existence, Multi-hop, and Negation, each posing unique challenges. For evaluation, we randomly sample 2K claims from the test set, ensuring balanced representation across categories under computational efficiency. To assess the generalizability of ClaimPKG beyond structured benchmarks, we also evaluate HoVer \cite{hover} and FEVEROUS \cite{feverous}, two widely-used unstructured-based benchmarks requiring multi-hop reasoning and evidence aggregation from Wikipedia. Additional statistics of datasets are provided in Appendix \ref{dataset_statistic}.
\\
\textbf{Metrics.} We use \textit{Accuracy} as the primary metric along with \textit{Entity Correctness} to measure if the claim's extracted entity is valid in KG. Additionally, for the FactKG dev set, we report \textit{Claim Structure Coverage}, which quantifies the proportion of triplets from the original claim's graph structure successfully reconstructed by our pipeline. We refer readers to Appendix \ref{appendix:additional_experiments} for more details.
\\
\textbf{Annotation.} For brevity, we use Llama-3B, Llama-70B, and Qwen-72B to refer to Llama-3.2-3B, Llama-3.3-70B, and Qwen2.5-72B respectively. The * symbol denotes models fine-tuned for pseudo subgraph generation. Full model names are used when necessary.
\\
\textbf{Baselines.} We compare ClaimPKG with recent KG-based claim verification methods: \textbf{Zero-shot CoT} \cite{chainofthought} prompts LLMs to generate rationales and verdicts without accessing the KG; \textbf{GEAR} \cite{gear}, originally designed for text-based verification, employs graph-based evidence aggregation with multiple aggregators to capture multi-evidence dependencies, using BERT for language representation and adapted for KG settings following \cite{factkg}; and \textbf{KG-GPT} \cite{kg_gpt}, a pioneer work that combines LLMs and KGs through a structured pipeline of Sentence Segmentation, Graph Retrieval, and Logic Inference. Notably, unlike baselines which receive pre-identified claim entities along with the claim as the input, our method processes entities in an end-to-end pipeline.
\\
\textbf{Implementation.} For a comprehensive evaluation, we evaluate baselines on three model series: Llama 3 \cite{llama3}, Qwen 2.5 \cite{qwen2.5}, and GPT-4o-mini \cite{gpt-4o-mini}. In ClaimPKG, we configure the Specialized LLM to generate multiple pseudo-subgraphs using a beam size of 5. For the Subgraph Retrieval algorithm, we adopt an embedding-based approach leveraging BGE-Large-EN-v1.5 \cite{bge_embedding} to compute dot-product similarity for the Relation Scoring Function, we set the primary hyperparameters to $k_1=3$ and $k_2=1$. Detailed justification is provided in Appendix \ref{appendix:additional_experiments}.
\subsection{Results and Analysis}

\begin{table*}[!h]
\centering
\label{table:performance}
\begin{adjustbox}{width=1.0\textwidth}
\begin{tabular}{l|c|ccccc|c}
\toprule
\multicolumn{1}{c|}{\textbf{Method}} & \textbf{Entity Correctness} & \textbf{Negation} & \textbf{Existence} & \textbf{Conjunction} & \textbf{Multi-hop} & \textbf{One-hop} & \textbf{Average} \\
\midrule
\rowcolor{lightgray} 
\multicolumn{8}{c}{Direct Inference With CoT - w/o Evidence Retrieval} \\
\midrule
GPT-4o-mini (Zero-shot CoT)                            &-              & 61.91 & 59.45 & 69.51 & 60.87 & 70.83 & 64.51 \\
Qwen-72B (Zero-shot CoT)                        &-              & 62.91 & 62.20 & 74.04 & 62.32 & 75.98 & 67.49 \\
Llama-70B (Zero-shot CoT)                       &-              & 64.34 & 64.62 & 72.47 & 65.58 & 78.32 & 69.07 \\
\midrule
\rowcolor{lightgray} 
\multicolumn{8}{c}{Baseline Comparision - w/ Evidence Retrieval} \\
\midrule
GEAR (Finetuned BERT)                            & Known in Prior & 79.72 & 79.19 & 78.63 & 68.39 & 77.34 & 76.65 \\
KG-GPT (Llama-70B Few-shot)                & Known in Prior & 70.91 & 65.06 & 86.64 & 58.87 & \textbf{92.02} & 74.70 \\
KG-GPT (Qwen-72B Few-shot)                  & Known in Prior & 67.31 & 60.08 & \textbf{89.14} & 58.19 & 90.87 & 73.12 \\
\textbf{ClaimPKG} (Llama-3B$^*$ + GPT-4o-mini)      & 100.0\%       & 85.10 & 72.64 & 84.23 & 72.26 & 91.01 & 81.05 \\
\textbf{ClaimPKG} (Llama-3B$^*$ + Qwen-72B)    & 100.0\%       & \textbf{85.27} & \textbf{86.90} & 84.02 & \textbf{78.71} & 91.20 & \textbf{85.22} \\
\textbf{ClaimPKG} (Llama-3B$^*$ + Llama-70B)   & 100.0\%       & 84.58 & 84.20 & 85.68 & 78.49 & 90.26 & 84.64 \\
\midrule
\rowcolor{lightgray} 
\multicolumn{8}{c}{Ablation Results (Llama-3B$^*$ + Llama-70B) - w/ Evidence Retrieval} \\
\midrule
ClaimPKG (w/o Trie Constraint)               & 87.50\%       & 82.50 & 83.24 & 83.82 & 76.13 & 88.01 & 82.74 \\
ClaimPKG (Few-shot Specialized LLM)         & 86.52\%       & 77.99 & 81.89 & 77.80 & 68.82 & 81.65 & 77.63 \\
ClaimPKG (w/o Incomplete Retrieval)           & 100.0\%       & 68.80 & 51.25 & 67.84 & 61.29 & 76.22 & 65.08 \\
\bottomrule
\end{tabular}
\end{adjustbox}
\caption{Performance (accuracy \%) comparison of ClaimPKG with baselines on 5 claim categories of FactKG dataset and their average scores.}
\label{tab:main_table}
\end{table*}
We present the main experimental results in this section and additional findings in Appendix~\ref{appendix:additional_experiments}.
\\[0.1cm]
\textbf{(RQ1): How Does ClaimPKG Perform Against the Baselines?}  
Table~\ref{tab:main_table} compares the accuracy (\%) of ClaimPKG with baselines across claim categories of the FactKG. Key observations include:  
\\
\textbf{(1)} Direct inference using LLMs with CoT reasoning significantly underperforms compared to evidence-based methods, with the best average score reaching only 69.07\%, highlighting that despite LLM advancements, evidence retrieval remains crucial. \textbf{(2)} KG-GPT integrates knowledge graphs with LLMs but its best average score achieves only 74.70\% (Llama-70B Few-shot), falling short of GEAR's fine-tuned model at 76.65\%. This suggests that while LLMs excel at language tasks, they require specific adaptation for KG processing.
\textbf{(3)} ClaimPKG, with the strongest configuration (Llama-3B$^*$ + Llama-70B) and constrained by Entity-Trie for valid KG entity generation, achieves a 12-point improvement over KG-GPT and 9 points over GEAR. It particularly excels in multi-hop reasoning, demonstrating strong performance across Llama-3 and Qwen-2.5 backbones through effective structured evidence retrieval and KG integration.
\\[0.1cm]
\textbf{(RQ2): How Do Different Components Affect Performance?}  
To evaluate the impact of each component in ClaimPKG, we conduct ablation studies of the following components, maintaining $\text{Llama-3B}^*$ as the Specialized LLM and Llama-70B as the General LLM.
\\
\textbf{Entity-Trie Constraint.} 
We remove the Entity-Trie constraint to assess its necessity. Compared to the full setup, this reduces the entity extraction correctness from 100\% to 87.5\%, and overall performance from 84.64\% to 82.72\%.
\\
\textbf{Specialized LLM.} 
When replacing the specialized LLM with few-shot prompting strategy using Llama-70B, a much larger general-purpose LLM, entity correctness further declines to 86.52\%, leading overall performance to drop to 77.63\%. These results demonstrate that even with examples, general-purpose LLMs struggle to produce outputs with desired graph structure correctly, emphasizing the importance of the specialized LLM in generating pseudo subgraphs.
\\
\textbf{Incomplete Retrieval.}  
Removing the Incomplete Triplet Retrieval function, which forces the retrieval algorithm to only query evidence using complete triplets, causes a significant average performance drop of nearly 20\% compared to the full setup, showing the complete graph structure of input claims is essential for optimal performance.
\\[0.1cm]
\textbf{(RQ3): Robustness and Generalization of ClaimPKG?}  
To assess ClaimPKG's robustness, we vary model backbones, examine zero-shot generalizability, analyze the effect of training data size, and conduct error analysis.
\\[0.08cm]
\textbf{Model Backbones.}  
We evaluate different LLM architectures for both Specialized and General LLMs (Table~\ref{tab:varying_baselines}). For General LLMs, we test various model sizes (7B to 70B parameters) using retrieved KG triplets as input. For Specialized LLMs, we experiment with different small fine-tuned backbones and few-shot prompt templates (Figure~\ref{fig:prompt_fewshot_pseudo_subgraph}), while keeping Llama-3.3-70B as the fixed General LLM.


\begin{table}[!h]

\centering

\label{table:performance}

\begin{adjustbox}{width=0.90\columnwidth}

\begin{tabular}{c|c|l|c}

\toprule

\textbf{Component} & \textbf{Strategy} & \multicolumn{1}{c|}{\textbf{Backbone}} & \textbf{Average} \\ \midrule

\multirow{6}{*}{\shortstack{General\\LLM}} & \multirow{5}{*}{Zero-shot} & Llama 3.1 - 8B & 77.08 \\

& & Llama 3.3 - 70B & 84.64 \\

& & GPT4o - Mini & 81.05 \\ 

& & Qwen 2.5 - 7B & 80.22 \\ 

& & Qwen 2.5 - 72B & 85.22 \\ \midrule

\multirow{6}{*}{\shortstack{Specialized\\LLM}} & \multirow{4}{*}{Finetune} & Llama 3 - 3B & 84.64 \\

& & Qwen 2.5 - 3B & 82.32 \\ 

& & Llama 3 - 1B & 83.91 \\

& & Qwen 2.5 - 1.5B & 82.20 \\ \cmidrule(){2-4}  

& \multirow{2}{*}{Few-shot} & Llama 3.3 - 70B & 77.63 \\

& & Qwen 2.5 - 72B & 77.10 \\ \bottomrule

\end{tabular}

\end{adjustbox}

\caption{Performance on Different Backbones.}

\label{tab:varying_baselines}

\end{table}  
Results in Table~\ref{tab:varying_baselines} show larger General LLMs (GPT-4o-Mini, Llama-3.3-70B) outperform smaller ones (Qwen-2.5-7B, Llama-3.1-8B) by up to 8 points, highlighting model capacity's role in aggregating subgraph evidence. Notably, a fine-tuned 1B Specialized LLM outperforms the general 70B counterpart, demonstrating fine-tuning's effectiveness to process graph data. This supports the need to combine powerful General LLMs with adapted Specialized LLMs for optimal performance.
\\[0.05cm]
\textbf{Zero-shot Generalizability.}  
\begin{table}[!h]
\centering
\label{table:performance}
\begin{adjustbox}{width=\columnwidth}
\begin{tabular}{l|c|c}
\toprule
\multicolumn{1}{c|}{\textbf{Benchmark}} & \textbf{Llama 3} & \textbf{Qwen 2.5} \\
\midrule
HoVer (Zero-shot CoT)                & 66.6           & 65.3           \\
HoVer (Support-Predicted)          & 70.7 (14.3\%) & 69.4 (15.7\%)	\\
\midrule
FEVEROUS (Zero-shot CoT)             & 81.1           & 80.9           \\
FEVEROUS (Support-Predicted)       & 83.8 (12.5\%)  & 83.6 (12.9\%)  \\
\bottomrule
\end{tabular}
\end{adjustbox}
\caption{Zero-shot transferred performance on other unstructure-based benchmarks on the Support-Predicted samples along with Support Predicted rates.}
\label{tab:generalization}
\end{table}
  
To assess ClaimPKG's zero-shot generalizability, we test transfer to HoVer~\cite{hover} and FEVEROUS~\cite{feverous} datasets. Using DBpedia~\cite{dbpedia} as the knowledge source, we evaluate with trained Specialized LLMs (Llama-3.2-3B and Qwen-2.5-3B) while keeping Llama-3.3-70B as the General LLM. Since external datasets may contain claims outside DBpedia's coverage, making it difficult to distinguish between knowledge gaps and actual verification failures of ClaimPKG for \textit{Refuted} cases, we analyze only samples predicted as \textit{Supported}. As shown in Table~\ref{tab:generalization}, ClaimPKG predicts \textit{Supported} for only 12.5\%-15.7\% of samples, indicating limited knowledge overlap with DBpedia. However, on these samples, ClaimPKG outperforms Llama-3.3-70B's zero-shot CoT inference by ~4\% accuracy on both datasets, demonstrating robust transfer to reasoning patterns in unseen data.
\\[0.05cm]
\textbf{Training Data Size.}  
\begin{figure}[!h]
  \centering
  \includegraphics[width=0.93\columnwidth]{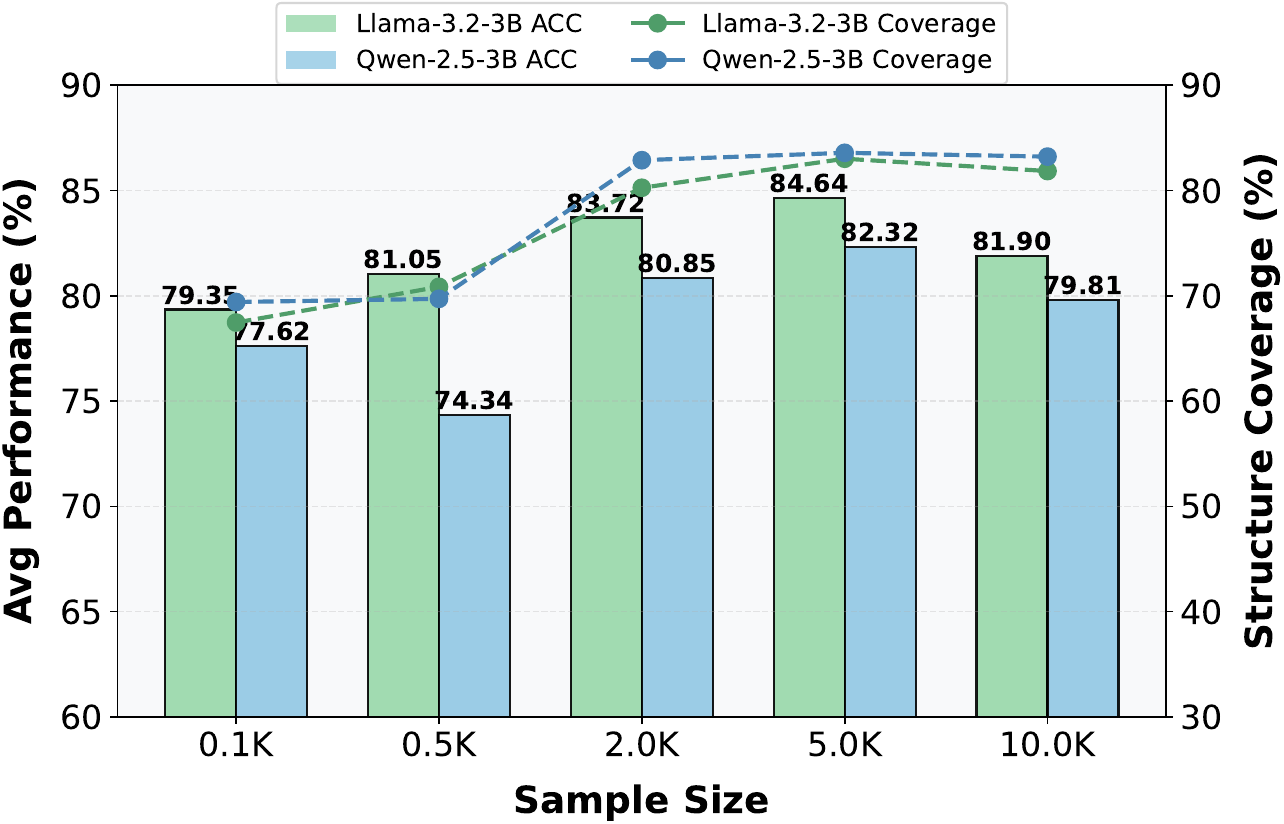}
  \caption{Varying Specialized LLM's training data.}
  \label{fig:varying_training_size}
\end{figure}  
To assess the impact of training data on the Specialized LLM, we vary the number of training samples from 0.1K to 10K, using two configurations: Llama-3.2-3B and Qwen-2.5-3B as the specialized LLM and keep the General LLM to be Llama-3.3-70B. We evaluate performance based on two metrics: average accuracy on the test set and claim structure coverage on the dev set. As shown in Figure~\ref{fig:varying_training_size}, the Specialized LLMs achieve satisfactory accuracy (Llama-3.2-3B: 79.35\%, Qwen-2.5-3B: 77.62\%) with just 100 training samples, demonstrating efficiency and low training costs for KG adaptation. While both structure coverage and accuracy improve up to 5K samples, coverage plateaus thereafter, and accuracy begins to decline, indicating overfitting where excessive training data reduces generalizability.

\subsection{Interpretability and Error Analysis}
\label{error_analysis}
ClaimPKG can improve claim verification performance while enhancing interpretability. Representative outputs of ClaimPKG (Figure \ref{tab:correct_samples}, Appendix \ref{appendix:case_study}) illustrate its ability to capture claim structure and provide well-grounded justifications. Notably, when refuting claims, it explicitly presents contradicting evidence, ensuring transparent reasoning. To further assess reliability, we conducted a human analysis of 200 incorrect predictions from FactKG, categorizing errors (Figure \ref{tab:incorrect_samples}, Appendix \ref{appendix:case_study}) into:
\textbf{Claim Structure Errors}: fail to capture the underlying claim structure;
\textbf{Retrieval Errors}: fail to retrieve necessary evidence required for claim verification; and
\textbf{Reasoning Errors}: incorrect logical inferences of the general LLM to judge the verdict.

Specifically, there are 0 (0\%) Claim Structure Errors, 57 (28.5\%) Retrieval Errors, and 143 (71.5\%) Reasoning Errors. These results suggest that, with chances (multiple beams) to generate pseudo-subgraphs, the Specialized LLM can effectively capture the structural representation of claims. However, the general-purpose LLM, despite its strong reasoning capabilities, still struggles with certain complex reasoning scenarios that require specific handling. Moreover, retrieval errors highlight cases where additional implicit reasoning is necessary, as we hypothesize that direct subgraph retrieval failed to provide a comprehensive picture of the required evidence. These highlight future improvements, focusing on enhancing retrieval inference and refining reasoning for complex claim verification over structured knowledge.
\subsection{Scalability of ClaimPKG}
ClaimPKG maintains scalability and adaptability within dynamic knowledge environments. After training the Specialized LLM on a domain (e.g., Wikipedia), the system remains decoupled from the underlying Knowledge Graph (KG). Only the Entity-Trie component interfaces directly with the data. Consequently, when the KG undergoes updates, ClaimPKG requires merely an update of the corresponding entities within the Entity-Trie, ensuring an efficient adaptation process.
\section{Conclusion}
In this work, we present ClaimPKG, a novel claim verification combining the structure of Knowledge Graphs with the adaptability and reasoning of Large Language Models. Through Pseudo-subgraph Generation, Subgraph Retrieval, and General Reasoning, it addresses limitations while ensuring transparency. Extensive experiments show state-of-the-art performance and generalizability across datasets, making ClaimPKG a step toward reliable and explainable misinformation detection.

\section*{Limitations}
Despite their advanced reasoning capabilities, LLMs are prone to errors and biases, necessitating careful deployment, particularly in fact-checking systems where incorrect or biased outputs could contribute to misinformation. Addressing these biases remains an ongoing research challenge, requiring effective mechanisms for detection, control, and mitigation. Additionally, real-world claim verification often requires inferring implicit reasoning, where further related knowledge for a problem is necessary, and making improvements in pipeline components to handle this type of information is crucial. Another limitation is the performance decline observed when the Specialized LLM is trained on an excessive number of examples, highlighting the need for future research into regularization strategies. Further improvements should also focus on the general reasoning module to infer missing knowledge more effectively and enhance intricate and nuanced claim verification cases over structured knowledge.


\bibliography{custom}

\begin{thebibliography}{34}
\providecommand{\natexlab}[1]{#1}

\bibitem[{Aly et~al.(2021)Aly, Guo, Schlichtkrull, Thorne, Vlachos, Christodoulopoulos, Cocarascu, and Mittal}]{feverous}
Rami Aly, Zhijiang Guo, Michael~Sejr Schlichtkrull, James Thorne, Andreas Vlachos, Christos Christodoulopoulos, Oana Cocarascu, and Arpit Mittal. 2021.
\newblock \href {https://datasets-benchmarks-proceedings.neurips.cc/paper/2021/hash/68d30a9594728bc39aa24be94b319d21-Abstract-round1.html} {{FEVEROUS:} fact extraction and verification over unstructured and structured information}.
\newblock In \emph{Proceedings of the Neural Information Processing Systems Track on Datasets and Benchmarks 1, NeurIPS Datasets and Benchmarks 2021, December 2021, virtual}.

\bibitem[{Cao et~al.(2021)Cao, Izacard, Riedel, and Petroni}]{autoregressiveentityretrieval}
Nicola~De Cao, Gautier Izacard, Sebastian Riedel, and Fabio Petroni. 2021.
\newblock \href {https://openreview.net/forum?id=5k8F6UU39V} {Autoregressive entity retrieval}.
\newblock In \emph{9th International Conference on Learning Representations, {ICLR} 2021, Virtual Event, Austria, May 3-7, 2021}. OpenReview.net.

\bibitem[{Edge et~al.(2024)Edge, Trinh, Cheng, Bradley, Chao, Mody, Truitt, and Larson}]{graphrag}
Darren Edge, Ha~Trinh, Newman Cheng, Joshua Bradley, Alex Chao, Apurva Mody, Steven Truitt, and Jonathan Larson. 2024.
\newblock \href {https://doi.org/10.48550/ARXIV.2404.16130} {From local to global: {A} graph {RAG} approach to query-focused summarization}.
\newblock \emph{CoRR}, abs/2404.16130.

\bibitem[{Glockner et~al.(2022{\natexlab{a}})Glockner, Hou, and Gurevych}]{task_definition_2}
Max Glockner, Yufang Hou, and Iryna Gurevych. 2022{\natexlab{a}}.
\newblock \href {https://doi.org/10.18653/V1/2022.EMNLP-MAIN.397} {Missing counter-evidence renders {NLP} fact-checking unrealistic for misinformation}.
\newblock In \emph{Proceedings of the 2022 Conference on Empirical Methods in Natural Language Processing, {EMNLP} 2022, Abu Dhabi, United Arab Emirates, December 7-11, 2022}, pages 5916--5936. Association for Computational Linguistics.

\bibitem[{Glockner et~al.(2022{\natexlab{b}})Glockner, Hou, and Gurevych}]{task_definition_3}
Max Glockner, Yufang Hou, and Iryna Gurevych. 2022{\natexlab{b}}.
\newblock \href {https://doi.org/10.18653/V1/2022.EMNLP-MAIN.397} {Missing counter-evidence renders {NLP} fact-checking unrealistic for misinformation}.
\newblock In \emph{Proceedings of the 2022 Conference on Empirical Methods in Natural Language Processing, {EMNLP} 2022, Abu Dhabi, United Arab Emirates, December 7-11, 2022}, pages 5916--5936. Association for Computational Linguistics.

\bibitem[{Herzig et~al.(2020)Herzig, Nowak, M{\"u}ller, Piccinno, and Eisenschlos}]{tapas}
Jonathan Herzig, Pawel~Krzysztof Nowak, Thomas M{\"u}ller, Francesco Piccinno, and Julian Eisenschlos. 2020.
\newblock \href {https://doi.org/10.18653/v1/2020.acl-main.398} {{T}a{P}as: Weakly supervised table parsing via pre-training}.
\newblock In \emph{Proceedings of the 58th Annual Meeting of the Association for Computational Linguistics}, pages 4320--4333, Online. Association for Computational Linguistics.

\bibitem[{Jiang et~al.(2023)Jiang, Zhou, Dong, Ye, Zhao, and Wen}]{struct-gpt}
Jinhao Jiang, Kun Zhou, Zican Dong, Keming Ye, Xin Zhao, and Ji-Rong Wen. 2023.
\newblock \href {https://doi.org/10.18653/v1/2023.emnlp-main.574} {{S}truct{GPT}: A general framework for large language model to reason over structured data}.
\newblock In \emph{Proceedings of the 2023 Conference on Empirical Methods in Natural Language Processing}, pages 9237--9251, Singapore. Association for Computational Linguistics.

\bibitem[{Jiang et~al.(2020)Jiang, Bordia, Zhong, Dognin, Singh, and Bansal}]{hover}
Yichen Jiang, Shikha Bordia, Zheng Zhong, Charles Dognin, Maneesh~Kumar Singh, and Mohit Bansal. 2020.
\newblock \href {https://doi.org/10.18653/V1/2020.FINDINGS-EMNLP.309} {Hover: {A} dataset for many-hop fact extraction and claim verification}.
\newblock In \emph{Findings of the Association for Computational Linguistics: {EMNLP} 2020, Online Event, 16-20 November 2020}, volume {EMNLP} 2020 of \emph{Findings of {ACL}}, pages 3441--3460. Association for Computational Linguistics.

\bibitem[{Kim et~al.(2023{\natexlab{a}})Kim, Kwon, Jo, and Choi}]{kg_gpt}
Jiho Kim, Yeonsu Kwon, Yohan Jo, and Edward Choi. 2023{\natexlab{a}}.
\newblock \href {https://doi.org/10.18653/V1/2023.FINDINGS-EMNLP.631} {{KG-GPT:} {A} general framework for reasoning on knowledge graphs using large language models}.
\newblock In \emph{Findings of the Association for Computational Linguistics: {EMNLP} 2023, Singapore, December 6-10, 2023}, pages 9410--9421. Association for Computational Linguistics.

\bibitem[{Kim et~al.(2023{\natexlab{b}})Kim, Park, Kwon, Jo, Thorne, and Choi}]{factkg}
Jiho Kim, Sungjin Park, Yeonsu Kwon, Yohan Jo, James Thorne, and Edward Choi. 2023{\natexlab{b}}.
\newblock \href {https://doi.org/10.18653/V1/2023.ACL-LONG.895} {Factkg: Fact verification via reasoning on knowledge graphs}.
\newblock In \emph{Proceedings of the 61st Annual Meeting of the Association for Computational Linguistics (Volume 1: Long Papers), {ACL} 2023, Toronto, Canada, July 9-14, 2023}, pages 16190--16206. Association for Computational Linguistics.

\bibitem[{Kwon et~al.(2023)Kwon, Li, Zhuang, Sheng, Zheng, Yu, Gonzalez, Zhang, and Stoica}]{vllm}
Woosuk Kwon, Zhuohan Li, Siyuan Zhuang, Ying Sheng, Lianmin Zheng, Cody~Hao Yu, Joseph~E. Gonzalez, Hao Zhang, and Ion Stoica. 2023.
\newblock Efficient memory management for large language model serving with pagedattention.
\newblock In \emph{Proceedings of the ACM SIGOPS 29th Symposium on Operating Systems Principles}.

\bibitem[{Lehmann et~al.(2015)Lehmann, Isele, Jakob, Jentzsch, Kontokostas, Mendes, Hellmann, Morsey, van Kleef, Auer, and Bizer}]{dbpedia}
Jens Lehmann, Robert Isele, Max Jakob, Anja Jentzsch, Dimitris Kontokostas, Pablo~N. Mendes, Sebastian Hellmann, Mohamed Morsey, Patrick van Kleef, S{\"{o}}ren Auer, and Christian Bizer. 2015.
\newblock \href {https://doi.org/10.3233/SW-140134} {Dbpedia - {A} large-scale, multilingual knowledge base extracted from wikipedia}.
\newblock \emph{Semantic Web}, 6(2):167--195.

\bibitem[{Li et~al.(2013)Li, Ji, and Huang}]{joint_extract_1}
Qi~Li, Heng Ji, and Liang Huang. 2013.
\newblock \href {https://aclanthology.org/P13-1008/} {Joint event extraction via structured prediction with global features}.
\newblock In \emph{Proceedings of the 51st Annual Meeting of the Association for Computational Linguistics, {ACL} 2013, 4-9 August 2013, Sofia, Bulgaria, Volume 1: Long Papers}, pages 73--82. The Association for Computer Linguistics.

\bibitem[{Loshchilov and Hutter(2019)}]{adamw}
Ilya Loshchilov and Frank Hutter. 2019.
\newblock \href {https://openreview.net/forum?id=Bkg6RiCqY7} {Decoupled weight decay regularization}.
\newblock In \emph{7th International Conference on Learning Representations, {ICLR} 2019, New Orleans, LA, USA, May 6-9, 2019}. OpenReview.net.

\bibitem[{Luo et~al.(2024)Luo, Li, Haffari, and Pan}]{reasoningongraph}
Linhao Luo, Yuan{-}Fang Li, Gholamreza Haffari, and Shirui Pan. 2024.
\newblock \href {https://openreview.net/forum?id=ZGNWW7xZ6Q} {Reasoning on graphs: Faithful and interpretable large language model reasoning}.
\newblock In \emph{The Twelfth International Conference on Learning Representations, {ICLR} 2024, Vienna, Austria, May 7-11, 2024}. OpenReview.net.

\bibitem[{Meta(2024)}]{llama3}
Meta. 2024.
\newblock \href {https://llama.meta.com/llama3/} {Build the future of ai with meta llama 3, 2024.}

\bibitem[{Miwa and Bansal(2016)}]{joint_extract_2}
Makoto Miwa and Mohit Bansal. 2016.
\newblock \href {https://doi.org/10.18653/v1/P16-1105} {End-to-end relation extraction using {LSTM}s on sequences and tree structures}.
\newblock In \emph{Proceedings of the 54th Annual Meeting of the Association for Computational Linguistics (Volume 1: Long Papers)}, pages 1105--1116, Berlin, Germany. Association for Computational Linguistics.

\bibitem[{OpenAI(2024)}]{gpt-4o-mini}
OpenAI. 2024.
\newblock \href {https://openai.com/index/hello-gpt-4o/} {Hello gpt-4o, 2024a.}

\bibitem[{Pan et~al.(2023)Pan, Wu, Lu, Luu, Wang, Kan, and Nakov}]{programfc}
Liangming Pan, Xiaobao Wu, Xinyuan Lu, Anh~Tuan Luu, William~Yang Wang, Min{-}Yen Kan, and Preslav Nakov. 2023.
\newblock \href {https://doi.org/10.18653/V1/2023.ACL-LONG.386} {Fact-checking complex claims with program-guided reasoning}.
\newblock In \emph{Proceedings of the 61st Annual Meeting of the Association for Computational Linguistics (Volume 1: Long Papers), {ACL} 2023, Toronto, Canada, July 9-14, 2023}, pages 6981--7004. Association for Computational Linguistics.

\bibitem[{Park et~al.(2022)Park, Min, Kang, Zettlemoyer, and Hajishirzi}]{faviq}
Jungsoo Park, Sewon Min, Jaewoo Kang, Luke Zettlemoyer, and Hannaneh Hajishirzi. 2022.
\newblock \href {https://doi.org/10.18653/v1/2022.acl-long.354} {{F}a{VIQ}: {FA}ct verification from information-seeking questions}.
\newblock In \emph{Proceedings of the 60th Annual Meeting of the Association for Computational Linguistics (Volume 1: Long Papers)}, pages 5154--5166, Dublin, Ireland. Association for Computational Linguistics.

\bibitem[{{Qwen}(2024)}]{qwen2.5}
{Qwen}. 2024.
\newblock \href {https://qwenlm.github.io/blog/qwen2.5/} {Qwen2.5: A party of foundation models}.

\bibitem[{Schuster et~al.(2021)Schuster, Fisch, and Barzilay}]{vitamin-c}
Tal Schuster, Adam Fisch, and Regina Barzilay. 2021.
\newblock \href {https://doi.org/10.18653/v1/2021.naacl-main.52} {Get your vitamin {C}! robust fact verification with contrastive evidence}.
\newblock In \emph{Proceedings of the 2021 Conference of the North American Chapter of the Association for Computational Linguistics: Human Language Technologies}, pages 624--643, Online. Association for Computational Linguistics.

\bibitem[{Sun et~al.(2024)Sun, Xu, Tang, Wang, Lin, Gong, Ni, Shum, and Guo}]{thinkongraph}
Jiashuo Sun, Chengjin Xu, Lumingyuan Tang, Saizhuo Wang, Chen Lin, Yeyun Gong, Lionel~M. Ni, Heung{-}Yeung Shum, and Jian Guo. 2024.
\newblock \href {https://openreview.net/forum?id=nnVO1PvbTv} {Think-on-graph: Deep and responsible reasoning of large language model on knowledge graph}.
\newblock In \emph{The Twelfth International Conference on Learning Representations, {ICLR} 2024, Vienna, Austria, May 7-11, 2024}. OpenReview.net.

\bibitem[{Thorne and Vlachos(2018)}]{task_definition_1}
James Thorne and Andreas Vlachos. 2018.
\newblock \href {https://aclanthology.org/C18-1283/} {Automated fact checking: Task formulations, methods and future directions}.
\newblock In \emph{Proceedings of the 27th International Conference on Computational Linguistics, {COLING} 2018, Santa Fe, New Mexico, USA, August 20-26, 2018}, pages 3346--3359. Association for Computational Linguistics.

\bibitem[{Thorne et~al.(2018)Thorne, Vlachos, Christodoulopoulos, and Mittal}]{fever}
James Thorne, Andreas Vlachos, Christos Christodoulopoulos, and Arpit Mittal. 2018.
\newblock \href {https://doi.org/10.18653/V1/N18-1074} {{FEVER:} a large-scale dataset for fact extraction and verification}.
\newblock In \emph{Proceedings of the 2018 Conference of the North American Chapter of the Association for Computational Linguistics: Human Language Technologies, {NAACL-HLT} 2018, New Orleans, Louisiana, USA, June 1-6, 2018, Volume 1 (Long Papers)}, pages 809--819. Association for Computational Linguistics.

\bibitem[{Wang et~al.(2020)Wang, Shin, Liu, Polozov, and Richardson}]{rat-sql}
Bailin Wang, Richard Shin, Xiaodong Liu, Oleksandr Polozov, and Matthew Richardson. 2020.
\newblock \href {https://doi.org/10.18653/v1/2020.acl-main.677} {{RAT-SQL}: Relation-aware schema encoding and linking for text-to-{SQL} parsers}.
\newblock In \emph{Proceedings of the 58th Annual Meeting of the Association for Computational Linguistics}, pages 7567--7578, Online. Association for Computational Linguistics.

\bibitem[{Wang and Shu(2023)}]{folk}
Haoran Wang and Kai Shu. 2023.
\newblock \href {https://doi.org/10.18653/V1/2023.FINDINGS-EMNLP.416} {Explainable claim verification via knowledge-grounded reasoning with large language models}.
\newblock In \emph{Findings of the Association for Computational Linguistics: {EMNLP} 2023, Singapore, December 6-10, 2023}, pages 6288--6304. Association for Computational Linguistics.

\bibitem[{Wang et~al.(2023)Wang, Wei, Schuurmans, Le, Chi, Narang, Chowdhery, and Zhou}]{self_consistency}
Xuezhi Wang, Jason Wei, Dale Schuurmans, Quoc~V. Le, Ed~H. Chi, Sharan Narang, Aakanksha Chowdhery, and Denny Zhou. 2023.
\newblock \href {https://openreview.net/forum?id=1PL1NIMMrw} {Self-consistency improves chain of thought reasoning in language models}.
\newblock In \emph{The Eleventh International Conference on Learning Representations, {ICLR} 2023, Kigali, Rwanda, May 1-5, 2023}. OpenReview.net.

\bibitem[{Wei et~al.(2022)Wei, Wang, Schuurmans, Bosma, Ichter, Xia, Chi, Le, and Zhou}]{chainofthought}
Jason Wei, Xuezhi Wang, Dale Schuurmans, Maarten Bosma, Brian Ichter, Fei Xia, Ed~H. Chi, Quoc~V. Le, and Denny Zhou. 2022.
\newblock \href {http://papers.nips.cc/paper\_files/paper/2022/hash/9d5609613524ecf4f15af0f7b31abca4-Abstract-Conference.html} {Chain-of-thought prompting elicits reasoning in large language models}.
\newblock In \emph{Advances in Neural Information Processing Systems 35: Annual Conference on Neural Information Processing Systems 2022, NeurIPS 2022, New Orleans, LA, USA, November 28 - December 9, 2022}.

\bibitem[{{Wikipedia}(2025{\natexlab{a}})}]{levenshtein}
{Wikipedia}. 2025{\natexlab{a}}.
\newblock \href {https://en.wikipedia.org/wiki/Levenshtein_distance} {{Levenshtein distance --- Wikipedia, The Free Encyclopedia}}.
\newblock Accessed: 14-February-2025.

\bibitem[{{Wikipedia}(2025{\natexlab{b}})}]{trie}
{Wikipedia}. 2025{\natexlab{b}}.
\newblock \href {https://en.wikipedia.org/wiki/Trie} {{Trie --- Wikipedia{,} The Free Encyclopedia}}.
\newblock [Online; accessed 9-February-2025].

\bibitem[{Xiao et~al.(2023)Xiao, Liu, Zhang, and Muennighoff}]{bge_embedding}
Shitao Xiao, Zheng Liu, Peitian Zhang, and Niklas Muennighoff. 2023.
\newblock \href {https://arxiv.org/abs/2309.07597} {C-pack: Packaged resources to advance general chinese embedding}.
\newblock \emph{Preprint}, arXiv:2309.07597.

\bibitem[{Zhou et~al.(2020)Zhou, Cui, Hu, Zhang, Yang, Liu, Wang, Li, and Sun}]{graph-review}
Jie Zhou, Ganqu Cui, Shengding Hu, Zhengyan Zhang, Cheng Yang, Zhiyuan Liu, Lifeng Wang, Changcheng Li, and Maosong Sun. 2020.
\newblock \href {https://doi.org/10.1016/J.AIOPEN.2021.01.001} {Graph neural networks: {A} review of methods and applications}.
\newblock \emph{{AI} Open}, 1:57--81.

\bibitem[{Zhou et~al.(2019)Zhou, Han, Yang, Liu, Wang, Li, and Sun}]{gear}
Jie Zhou, Xu~Han, Cheng Yang, Zhiyuan Liu, Lifeng Wang, Changcheng Li, and Maosong Sun. 2019.
\newblock \href {https://doi.org/10.18653/V1/P19-1085} {{GEAR:} graph-based evidence aggregating and reasoning for fact verification}.
\newblock In \emph{Proceedings of the 57th Conference of the Association for Computational Linguistics, {ACL} 2019, Florence, Italy, July 28- August 2, 2019, Volume 1: Long Papers}, pages 892--901. Association for Computational Linguistics.

\end{thebibliography}
\appendix
\section{Benchmark Datasets}
\begin{table}[!h]
    \centering
    \begin{adjustbox}{width=0.85\columnwidth}
    \begin{tabular}{crrrrr}
        \toprule
        \toprule
        \multicolumn{1}{c}{\textbf{Dataset}} &
        \multicolumn{1}{c}{\textbf{Split}} &
        \multicolumn{1}{c}{\textbf{Support}} &
        \multicolumn{1}{c}{\textbf{Refute}} &
        \multicolumn{1}{c}{\textbf{NEI}} &
        \multicolumn{1}{c}{\textbf{Total}} \\
        \midrule
        \multirow{4}{*}{\shortstack{FactKG}}
          & Train & 42723 & 43644 & - & 86367 \\
          & Dev & 6426 & 6840 & - & 132666 \\
          & Test & 4398 & 4643 & - & 9041 \\
          & Total & 53547 & 55127 & - & 108674 \\
        \midrule
        \multirow{4}{*}{\shortstack{Hover}}
          & Train & 11023 & 7148 & - & 18171 \\
          & Dev & 2000 & 2000 & - & 4000 \\
          & Test & 2000 & 2000 & - & 4000 \\
          & Total & 15023 & 11148 & - & 26171 \\
        \midrule
        \multirow{4}{*}{\shortstack{FEVER\\OUS}}
          & Train & 41835 & 27215 & 2241 & 71291 \\
          & Dev & 3908 & 3481 & 501 & 7890 \\
          & Test & 3372 & 2973 & 1500 & 7845 \\
          & Total & 49115 & 33669 & 4242 & 87026 \\
        \bottomrule
        \bottomrule
    \end{tabular}
    \end{adjustbox}
    \caption{Basic statistics of Hover, FEVEROUS, and FactKG Datasets}
    \label{tab:combined_stats}
\end{table}
\begin{table}[!h]
    \centering
    \begin{adjustbox}{width=0.88\columnwidth}
    \begin{tabular}{crrrr}
        \toprule
        \toprule
        \multirow{2}{*}{\textbf{Type}} & \multirow{2}{*}{\textbf{Written}} & \multicolumn{2}{c}{\textbf{Colloquial}} & \multirow{2}{*}{\textbf{Total}} \\
        \cmidrule(lr){3-4}
         &  & \textbf{Model} & \textbf{Presup} &  \\
        \midrule
        One-hop & 2,106 & 15,934 & 1,580 & 19,530 \\
        Conjunction & 20,587 & 15,908 & 602 & 37,097 \\
        Existence & 280 & 4,060 & 4,832 & 9,172 \\
        Multi-hop & 10,239 & 16,420 & 603 & 27,262 \\
        Negation & 1,340 & 12,466 & 1,807 & 15,613 \\
        \midrule
        Total & 34,462 & 64,788 & 9,424 & 108,674 \\
        \bottomrule
        \bottomrule
    \end{tabular}
    \end{adjustbox}
    \caption{Dataset statistics of FACTKG for claim types.}
    \label{tab:factkg_stats}
\end{table}
\textbf{FEVEROUS.} \cite{feverous} FEVEROUS is a fact verification dataset comprising 87,026 verified claims sourced from Wikipedia (Table \ref{tab:combined_stats}). Each claim is accompanied by evidence in the form of sentences and/or cells from tables, along with a label indicating whether the evidence supports, refutes, or does not provide enough information to verify the claim. The dataset includes metadata like annotator actions and challenge types, designed to minimize biases. It is used for tasks that involve verifying claims against both unstructured (textual) and structured (tabular) information.
\\[0.1cm]
\textbf{HoVer.} \cite{hover} HoVer is a dataset containing 26,171 samples, designed for open-domain, multi-hop fact extraction and claim verification, using the Wikipedia corpus. Claims in HoVer are adapted from question-answer pairs and require the extraction of facts from multiple (up to four) Wikipedia articles to determine if the claim is supported or not supported. The complexity of HoVer, particularly in the 3/4-hop claims, is further amplified because these claims are often expressed across multiple sentences, which introduces challenges related to long-range dependencies, such as accurately resolving coreferences. 
\\[0.1cm]
\textbf{FactKG.} \cite{factkg} FactKG is a challenging fact verification dataset comprised of 108,674 samples, designed to rigorously test models' abilities to reason over structured knowledge represented in a knowledge graph. Its difficulty arises from a combination of factors. First, it demands proficiency in five distinct reasoning types: one-hop (single relationship), conjunction (combining multiple relationships), existence (verifying entity/relationship presence), multi-hop (traversing multiple relationships), and, crucially, negation (reasoning about the absence of relationships). Second, FactKG incorporates linguistic diversity, encompassing both formal, written-style claims and more challenging colloquial expressions, requiring models to handle paraphrasing, idiomatic language, and less direct wording. Third, instead of unstructured text, FactKG utilizes the DBpedia knowledge graph (derived from Wikipedia), necessitating that models correctly link entities and relations mentioned in the claim to the graph's nodes and edges, and perform complex path-based reasoning, especially for multi-hop claims. The addition of a weakly semantic knowledge source, and cross-style evaluation to asses generalizability, further contributes to the difficulty of this dataset. These features collectively make FactKG significantly more complex than datasets relying solely on unstructured text for verification. Detailed statistics of this dataset can be found in table \ref{tab:factkg_stats}.
Readers can refer to table \ref{tab:combined_stats} for the overall basic statistics of all employed datasets for ClaimPKG.
\label{dataset_statistic}
\section{Implementation Details}
We conducted all experiments on a DGX server with 8 NVIDIA A100 GPUs. The General LLM is hosted within the vLLM framework \cite{vllm}. Below, we detail the training process of the Specialized LLM.
\subsection{Specialized LLM Training Data Annotation}
\label{training_data_annotation}
To tailor the specialized model for improved comprehension and processing of KG-specific data, we construct a dedicated dataset for training, leveraging the provided version of FactKG~\cite{factkg} (illustrated in Figure~\ref{fig:factkg_original_data_example}). The annotation process consists of the following steps:
\begin{figure}[htbp]
    \centering
    \begin{tcolorbox}[
        colback=gray!10,
        boxrule=0.4pt,
        arc=2pt,
        left=2pt,     
        right=2pt,    
        top=2pt,       
        bottom=2pt,     
        fontupper=\small
    ]
    \textbf{Claim:} A musical artist, whose music is Post-metal, played with the band Twilight and performs for Mamiffer. \\
    \textbf{Entities:} [Mamiffer, Post-metal, Twilight\_(band)] \\
    \textbf{Evidence:} \\
    - Twilight\_(band), (~associatedMusicalArtist, associatedBand), Mamiffer) \\
    - Twilight\_(band), (~associatedMusicalArtist, genre), Post-metal
    \end{tcolorbox}
    \caption{Provided data of FactKG}
    \label{fig:factkg_original_data_example}
\end{figure}

\paragraph{Preprocessing:} All entities and relations from FactKG, including the train, development, and test datasets, as well as the DBPedia KG, are normalized by splitting concatenated words to ensure consistency.
\paragraph{Graph Construction:} Using the provided evidence information from FactKG, we observe that while evidence may not explicitly exist in the graph, it accurately captures the underlying structure of the claim. Accordingly, for triplets with relation paths exceeding one hop, we decompose them into multiple triplets while introducing a placeholder entity, denoted as \texttt{``unknown\_\{index\}''}, to preserve structural integrity. This placeholder represents an ambiguous or missing entity that requires identification.  For instance, the triplet: \texttt{``Twilight\_(band), ($\sim$associatedMusicalArtist, associatedBand), Mamiffer''} is transformed into the following triplets: \texttt{``Twilight\_(band), associatedBand, unknown\_1''} and \texttt{``unknown\_1, associatedMusicalArtist, Mamiffer''}.
Additionally, entities present in the \textbf{Entities} set but absent from the graph are also introduced as \texttt{unknown\_\{index\}}. To further enhance graph completeness, GPT-4 is employed to verify whether entities from the \textbf{Entities} set are explicitly mentioned in the claim. This ensures that relevant entities are either linked to existing nodes or added as placeholders. The automatic entity verification process is conducted using a prompt template, as shown in Figure~\ref{fig:prompt_in_out_entities}. Additionally, the symbol "\textasciitilde{}" is retained to denote inverse relations. Random shuffle among constructed triplets but preserving the sequential order of \texttt{``unknown''} entity is applied to improve the robustness of the model being trained.
\paragraph{Generated Pseudo-Subgraph:} The transformed claim results in the pseudo-subgraph illustrated in Figure \ref{fig:pseudo_subgraph_label}.

\begin{figure}[htbp]
    \centering
    \begin{tcolorbox}[
        colback=gray!10,
        boxrule=0.4pt,
        arc=2pt,
        left=2pt,     
        right=2pt,    
        top=2pt,       
        bottom=2pt,     
        fontupper=\small
    ]
    \textbf{Pseudo Subgraph Label:} \\
    - Twilight (band), ~associated musical artist, unknown\_0 \\
    - unknown\_0, associated band , Mamiffer \\
    - unknown\_0, genre, Post-metal
    \end{tcolorbox}
    \caption{Pseudo-Subgraph label as the output of the data annotation process.}
    \label{fig:pseudo_subgraph_label}
\end{figure}


\subsection{Training and Hyperparameter Settings of the Specialized LLM}
\begin{table}[h]
    \centering
    \begin{adjustbox}{width=0.7\columnwidth}
    \begin{tabular}{l|c}
        \toprule
        \toprule
        \textbf{Parameter} & \textbf{Value} \\
        \midrule
        \multirow{2}{*}{\textbf{Backbone}} & Llama-3-Base \\
        & Qwen-2.5-Base \\
        \textbf{Learning Rate} & 1e-5 \\
        \textbf{Training Epoch} & 1 \\
        \textbf{Training Steps} & 128 \\
        \textbf{Optimizer} & AdamW \\
        \bottomrule
        \bottomrule
    \end{tabular}
    \end{adjustbox}
    \caption{Hyperparameters of the Specialized LLM in ClaimPKG.}
    \label{tab:training_config}
\end{table}
The training configurations for the Specialized LLM are summarized in Table \ref{tab:training_config}. The model training is based on the \textbf{Base} version of Llama-3 (Llama-3.2-1B, Llama-3.2-3B, Llama-3.1-8B) and Qwen 2.5 (Qwen-2.5-1.5B, Qwen-2.5-3B, Qwen-2.5-7B). These base models are selected to preserve their inherent linguistic capabilities while facilitating optimal adaptation to domain-specific tasks during fine-tuning. The training process employs the annotated dataset described in Section \ref{training_data_annotation} and is conducted over one single epoch using the AdamW \cite{adamw} optimizer. This strategy enables the generation of multiple variants of the Specialized LLM, ensuring task-specific adaptation while maintaining robust generalization across diverse linguistic structures.

\section{Additional Experimental Results}
In this section, we present additional experimental results through a systematic analysis on the FactKG development set with 2000 randomly sampled data points across claim categories. First, we provide a more detailed explanation of the evaluation metrics used. Second, we examine the performance of the specialized LLM by varying the beam size and backbone model size. Third, we analyze the Subgraph Retrieval by adjusting the hyperparameters $k_1$ and $k_2$ as explained in the \ref{section:subgraph_retrieval}, which influence the diversity and correctness of the retrieved subgraphs.
\label{appendix:additional_experiments}
\subsection{Metrics}
The specialized LLM's generation of pseudo-subgraphs plays a crucial role in ClaimPKG's performance. We evaluated the specialized LLM's performance using four metrics: claim structure coverage ($coverage$), entity correctness ($correctness$), unique triplet count, and average end-to-end accuracy. While the final metric is straightforward, the three former metrics can be described as follows:
\\[0.1cm]
\textit{ (1) Structure coverage} quantifies the alignment between the LLM-generated pseudo-graph and the reference claim graph in the FactKG dataset. Specifically, for a generated graph $P$ and reference graph $Q$, $coverage$ is computed as:
\begin{equation*}
    coverage(P,Q) = \frac{\#(P.triplets\ \cap\ Q.triplets)}{\#(Q.triplets)}
\end{equation*}
\textit{(2) Entity correctness} quantifies the correctness of a claim’s extracted entities, i.e., whether these entities exist in the KG. Specifically, for a generated graph $P$ and a knowledge graph $\mathcal{G}$, $correctness$ is computed as:
\begin{equation*}
    correctness(P,\mathcal{G}) = \frac{\#(P.enities\ \cap \mathcal{G}.entities)}{\#(P.entities)}
\end{equation*}
\textit{(3) Unique triplet count} measures the diversity of generated graph structures, with higher counts potentially enabling better subgraph retrieval through increased coverage of possible relationships.

\subsection{Different Beam Sizes of the Specialized LLM}

\begin{table}[!h]
    \centering
    \begin{adjustbox}{width=\columnwidth}
    \begin{tabular}{c|c|c|c|c}
        \toprule
        \toprule
        \textbf{Backbone} & \textbf{\shortstack{Beam\\Size}} & \textbf{\shortstack{Average\\Accuracy}} & \textbf{\shortstack{Structure\\Coverage}} & \textbf{\shortstack{Unique\\Triplets}} \\
        \midrule
        \multirow{4}{*}{Llama-3B}
          & Beam 1  & 79.78 & 76.51 & 4.48 \\
          & Beam 3  & 81.80 & 81.27 & 6.44 \\
          & Beam 5  & 82.04 & 83.02 & 8.39 \\
          & Beam 10 & 82.33 & 84.61 & 13.83 \\
        \midrule
        \multirow{4}{*}{Qwen-3B}
          & Beam 1  & 78.84 & 77.95 & 3.82 \\
          & Beam 3  & 80.76 & 82.66 & 5.16 \\
          & Beam 5  & 81.41 & 83.58 & 6.73 \\
          & Beam 10 & 82.19 & 84.62 & 9.58 \\
        \bottomrule
        \bottomrule
    \end{tabular}
    \end{adjustbox}
    \caption{Performance metrics for different models on FactKG dev set.}
    \label{tab:model_performance}
\end{table}
\begin{table}[h!]
    \centering
    \begin{adjustbox}{width=\columnwidth}
    \begin{tabular}{l|ccc}
        \toprule
        \toprule
        \textbf{Beam Size} & \textbf{Gen Graph (s)} & \textbf{Retrieve (s)} & \textbf{Reason (s)}\\
        \midrule
        beam 1  & 1.02  & 0.24 & 2.19 \\
        beam 3  & 2.16  & 0.38 & 2.22 \\
        beam 5  & 3.52  & 0.50 & 2.33 \\
        beam 10 & 35.18 & 1.01 & 2.88 \\
        \bottomrule
        \bottomrule
    \end{tabular}
    \end{adjustbox}
    \caption{Computing time for different beam sizes on FactKG dev set.}
    \label{tab:beam_metrics}
\end{table}
To evaluate the LLM's decoding strategy across different beam sizes, we utilized three average \textit{accuracy, structure coverage} and \textit{unique triplet count} as metrics. Table \ref{tab:model_performance} details the impact of the number of beam sizes on the previously mentioned metrics on the FactKG dev set. Both Llama and Qwen models demonstrate consistent improvements in average performance and claim structure coverage as beam size increases from 1 to 10. At beam size 10, Llama achieves 84.61\% coverage while Qwen reaches 84.62\%, showing comparable performance at higher beam sizes. The unique triplet count shows more pronounced growth with larger beam sizes, with Llama generating 13.83 unique triplets and Qwen 9.58 triplets at beam size 10.

However, table \ref{tab:beam_metrics} shows this improved performance comes with significant computational overhead. Table \ref{tab:beam_metrics} details on the time taken for generating pseudo-graphs, retrieving sub-graphs and reasoning with retrieved evidence. Most notably, while the time required for retrieving sub-graphs and reasoning with evidence only increase marginally as the beam size increase, this figure for pseudo-graph generation increases dramatically as the beam size goes to 10, from 1.02s at beam size 1 to 35.18s at beam size 10 - a 34.5× increase. Based on this measurement, in our official framework we select beam size = 5 to balance the performance gain and computational costs.

\subsection{Different Model Sizes of the Specialized LLM}
To evaluate how model size affects performance, we compare different variants of Llama and Qwen models ranging from 1B to 8B parameters. Table \ref{tab:ablation-model-sizes} presents the performance on the FactKG dev set across three key metrics: average performance, structure coverage, and unique triplets generated, which was explained previously.

\begin{table}[!h]
    \centering
    \begin{adjustbox}{width=0.9\columnwidth}
    \begin{tabular}{c|c|c|c}
        \toprule
        \toprule
        \textbf{Backbone} & \textbf{\shortstack{Average\\Accuracy}} & \textbf{\shortstack{Structure\\Coverage}} & \textbf{\shortstack{Unique\\Triplets}} \\
        \midrule
        Llama - 1B  & 80.26 & 78.98 & 8.97 \\
        Llama - 3B  & 82.04 & 83.02 & 8.39 \\
        Llama - 8B  & 82.63 & 82.84 & 9.34 \\
        \midrule
        Qwen - 1.5B & 80.48 & 81.34 & 6.58 \\
        Qwen - 3B   & 81.41 & 83.58 & 6.73 \\
        Qwen - 7B   & 81.79 & 82.88 & 7.05 \\
        \bottomrule
        \bottomrule
    \end{tabular}
    \end{adjustbox}
    \caption{Performance metrics for different models on the FactKG dev set.}
    \label{tab:ablation-model-sizes}
\end{table}
\
For both model families, we observe improvements in performance as model size increases, though with different patterns. The Llama family shows more notable gains, with average performance increasing from 80.26\% (1B) to 82.63\% (8B), while Qwen demonstrates more modest improvements from 80.48\% (1.5B) to 81.79\% (7B). Structure coverage peaks with the 3B variants for both families - Llama-3B achieving 83.02\% and Qwen-3B reaching 83.58\%. The models keep the increasing trend in their triplet generation patterns: Llama maintains relatively stable unique triplet counts (8.39 - 9.34) across sizes, while the figures for Qwen are (6.58 - 7.05) as the model size increases.

Overall, scaling to larger models shows slight improvements while increasing computational requirements. Based on these results, we select 3B variants of both model families in our official implementation, which offer an optimal balance of performance and model size, with Llama-3B and Qwen-3B showing comparable effectiveness across all metrics. 

\subsection{Different Hyperparameters of Subgraph Retrieval}
\begin{table}[!h]
    \centering
    \begin{adjustbox}{width=0.7\columnwidth}
    \begin{tabular}{c|c|c}
        \toprule
        \toprule
        \textbf{Hyper Params} & \textbf{\shortstack{Average\\Accuracy}} & \textbf{\shortstack{Unique\\Triplets}} \\
        \midrule
        $k_1=5; k_2=3$ & 82.00 & 11.42 \\
        $k_1=3; k_2=1$ & 82.04 & 8.39 \\
        $k_1=1; k_2=1$ & 81.87 & 3.58 \\
        \bottomrule
        \bottomrule
    \end{tabular}
    \end{adjustbox}
    \caption{Performance of different subgraph retrieval configurations $k_1$ and $k_2$ with Llama-3.2-3B + Llama-3.3-70B on the FactKG dev set.}
    \label{tab:different_retrieval_params}
\end{table}
To assess the impact of different hyperparameters in the subgraph retrieval algorithm on overall performance, we systematically vary these hyperparameters while keeping the specialized LLM and general LLM fixed as Llama-3.2-3B and Llama-3.3-70B, respectively. Table~\ref{tab:different_retrieval_params} presents the performance across two key metrics: average accuracy and the number of unique triplets generated. 

The results indicate that increasing $k_1$ and $k_2$ leads to a higher number of unique triplets, suggesting greater diversity in retrieved claims. However, this increase does not consistently translate to overall performance gains, which fall in the range of 81.87 - 82.00. Notably, performance peaks at $k_1 = 3$ and $k_2 = 1$, suggesting that a more focused retrieval strategy is sufficient to achieve optimal performance, whereas excessively high $k$ values may introduce noise or irrelevant information. Based on these results, we select $k_1 = 3$ and $k_2 = 1$ in our official implementation, which balancing between information discovery and computing required.

\subsection{Different Methods for Relation Scoring Function}
\begin{table}[h]
    \centering
    \begin{adjustbox}{width=0.8\columnwidth}
    \begin{tabular}{l|c}
        \toprule
        \toprule
        \textbf{Method} & \textbf{Average Accuracy} \\
        \hline
        Embedding Based             & 84.64 \\
        Rerank Based                & 84.73 \\
        Fuzzy Matching              & 82.19 \\
        Exact Matchching            & 81.57 \\
        \bottomrule
        \bottomrule
    \end{tabular}
    \end{adjustbox}
    \caption{Performance of different scoring approach of the Subgraph Retrieval on the FactKG test set}
    \label{tab:different_retrieval_methods}
\end{table}

To assess the impact of different scoring mechanisms on performance, we vary the scoring function and evaluate the test set of FactKG while fix the Specialized LLM and the General LLM. Specifically, we explore multiple strategies for the Relation Scoring Function (\textit{Sim}), as described in Section~\ref{section:subgraph_retrieval}, incorporating diverse techniques such as embedding-based retrieval, reranking, fuzzy text matching \cite{levenshtein}, and exact matching. 

For embedding-based and reranking approaches, we employ state-of-the-art pre-trained models, namely BGE-Large-EN-v1.5\footnote{https://huggingface.co/BAAI/bge-large-en-v1.5} and BGE-Reranker-Large\footnote{https://huggingface.co/BAAI/bge-reranker-large}, as provided by \cite{bge_embedding}. Experimental results indicate that deep learning-based methods, such as embedding and reranking, achieve superior performance, with accuracy scores of 84.64 and 84.56, respectively. In contrast, text-matching-based methods yield lower accuracy, with fuzzy matching and exact matching scoring 82.19 and 81.57, respectively. These findings highlight the effectiveness of deep learning-based approaches.

We recommend embedding-based retrieval as it enables pre-indexing of corpus relations. This allows precomputation of relation embeddings and requires encoding only the query relation for new Pseudo Subgraphs, eliminating the need to re-encode existing knowledge graph relations during inference.

\section{Algorithm Details}
\label{appendix:trie_constrained_decoding_details}
\begin{algorithm*}[h]
    \small
    \caption{LLM Decoding with Entity-Trie Constraint}
    \SetKwFunction{ConstrainedDecoding}{ConstrainedDecoding}
    \SetKwFunction{InitializeHiddenStates}{InitializeHiddenStates}
    \SetKwFunction{UpdateHiddenState}{UpdateHiddenState}
    \SetKwFunction{ExtractPrefix}{ExtractPrefix}
    \SetKwFunction{MaskProb}{MaskProb}
    \SetKwFunction{lookup}{lookup}
    \SetKwProg{Fn}{Function}{:}{}
    \SetKwInOut{Input}{Input}
    \SetKwInOut{Output}{Output}
    \Input{Specialized $LLM$, Input claim $c$, Entity Trie $\mathcal{T}$}
    \Output{Pseudo-Subgraph $\mathcal{P}$}

    \KwSty{Initialize:}\\
    \hspace{3mm} $\mathcal{P} \gets \emptyset$ \tcp*[r]{Initialize pseudo subgraph}
    \hspace{3mm} $h_0 \gets \InitializeHiddenStates()$\;
    \hspace{3mm} $constrained \gets \text{False}$\;
    \Fn{\ConstrainedDecoding{$LLM, c, \mathcal{T}$}}{
        \While{True}{
            $p_t, h_t \gets LLM(\mathcal{P}, c, h_{t-1})$ \tcp*[r]{Compute token probabilities and update hidden states}
            
            \If{$constrained$}{
                $prefix \gets \ExtractPrefix(\mathcal{P})$ \tcp*[r]{Retrieve tokens from last unclosed <e> to the last}
                $allowed \gets \mathcal{T}.\lookup(prefix)$ \tcp*[r]{Retrieve allowed tokens from valid continuations in $\mathcal{T}$}
                $p_t \gets \MaskProb(p_t, allowed)$ \tcp*[r]{Impose probabilities of invalid tokens to be 0}
            }
            
            $new\_token \gets \arg\max p_t$ \tcp*[r]{Select new token for $\mathcal{P}$}
            $\mathcal{P} \gets \mathcal{P} \cup \{new\_token\}$\;
            
            \If{$new\_token == \text{<e>}$}{
                $constrained \gets \text{True}$;\
            }
            \If{$new\_token == \text{</e>}$}{
                $constrained \gets \text{False}$;\
            }
            
            
            \If{$new\_token == \text{EOS}$}{
                \textbf{break};\
            }
        }
        \Return{$\mathcal{P}$}
    }
\label{algo:constrained_decoding}
\end{algorithm*}
The detailed implementation of the Entity Trie-constrained decoding algorithm is provided as the pseudo-code in Algorithm \ref{algo:constrained_decoding} and the Algorithm~\ref{algo:subgraph_retrieval} details the implementation of the Subgraph Retrieval.
\label{appendix:subgraph_retrieval_algorithm_details}

\begin{algorithm*}[h]
\small
\caption{Subgraph Retrieval}
\SetKwFunction{SubgraphRetrieval}{SubgraphRetrieval}
\SetKwFunction{RetrieveSingleSubgraph}{RetrieveSingleSubgraph}
\SetKwFunction{RetrieveIncomplete}{RetrieveIncomplete}
\SetKwFunction{RetrieveComplete}{RetrieveComplete}
\SetKwFunction{CategorizeTriplets}{CategorizeTriplets}
\SetKwFunction{JoinSubgraphs}{JoinSubgraphs}
\SetKwFunction{GroupByUnknown}{GroupTripletsByUnknown}
\SetKwFunction{ExtractStructure}{ExtractPseudoStructure}
\SetKwFunction{GetCandScores}{GetCandidatesAndScores}
\SetKwFunction{RankTopK}{RankTopKCandidates}
\SetKwFunction{GetTriplets}{GetTriplets}
\SetKwFunction{ExistConn}{ExistConnection}
\SetKwFunction{GetTripletsHT}{GetTripletsWithHeadAndTail}
\SetKwFunction{TopK}{TopK}
\SetKwFunction{RetrieveConnRels}{RetrieveActualConnectedRelations}
\SetKwFunction{RetrieveConnEnts}{RetrieveActualConnectedEntities}
\SetKwFunction{RelScore}{RelationScore}
\SetKwFunction{MaxRelScore}{MaxRelatedRelationScores}
\SetKwFunction{TopKCand}{TopKCandidates}
\SetKwFunction{AggregateGlobalScore}{AggregateGlobalScore}
\SetKwFunction{ExtractUnknownCandidates}{ExtractUnknownCandidates}
\SetKwFunction{Sum}{Sum}
\SetKwProg{Fn}{Function}{:}{}

\SetKwInOut{Input}{Input}
\SetKwInOut{Output}{Output}

\Input{Knowledge graph $\mathcal{G}$, Pseudo Subgraph List $\mathcal{P}_c$, Top $k_1$ Candidate Unknown Entities, Top $k_2$ Complete Triplets}
\Output{Combined subgraph $\mathcal{S}_c$}

\Fn{\SubgraphRetrieval{$\mathcal{G}, \mathcal{P}_c, k_1, k_2$}}{
    $S \gets \emptyset$\;
    \ForEach{$\mathcal{P}\in \mathcal{P}_c$}{
        $S \gets S \cup \RetrieveSingleSubgraph(\mathcal{G},\mathcal{P}, k_1, k_2)$ \tcp*[r]{Process each pseudo subgraph}
    }
    \Return{\JoinSubgraphs($S$)} \tcp*[r]{Combine subgraphs}
}

\Fn{\RetrieveSingleSubgraph{$\mathcal{G}, \mathcal{P}, k_1, k_2$}}{
    $(T_{\text{comp}},T_{\text{inc}}) \gets \CategorizeTriplets(\mathcal{P})$ \tcp*[r]{Split into complete/incomplete triplets}
    $S_{\text{inc}} \gets \RetrieveIncomplete(\mathcal{G}, T_{\text{inc}}, k_1)$\;
    $S_{\text{comp}} \gets \RetrieveComplete(\mathcal{G}, T_{\text{comp}}, k_1, k_2)$\;
    \Return{$S_{\text{inc}} \cup S_{\text{comp}}$}
}

\Fn{\RetrieveIncomplete{$\mathcal{G}, T_{\text{inc}}, k_1$}}{
    $S \gets \emptyset$\;
    $G \gets \GroupByUnknown(T_{\text{inc}})$ \tcp*[r]{Group by unknown entity}
    \ForEach{$g \in G$}{
        $(E_u,R_u) \gets \ExtractStructure(g)$ \tcp*[r]{Extract entities and relations associated to unknown entity}
        $C \gets \emptyset$\;
        \ForEach{$(e, r) \in (E_u,R_u)$}{
            $(C_e,\text{scores}) \gets \GetCandScores(\mathcal{G}, e, r)$\;
            $C \gets C \cup \{(C_e,\text{scores})\}$\;
        }
        $C = \AggregateGlobalScore(C)$ \tcp*[r]{Aggregate candidate scores globally}
        $C^* \gets \RankTopK(C, k_1)$ \tcp*[r]{Select top-$k_1$ candidates}
        $S \gets S \cup \GetTriplets(C^*, g)$\;
    }
    \Return{$S$}
}

\Fn{\GetCandScores{$\mathcal{G}, e, r$}}{
    $R_{\text{act}} \gets \RetrieveConnRels(\mathcal{G}, e)$\;
    $E_{\text{act}} \gets \RetrieveConnEnts(\mathcal{G}, e)$\;
    $r\_scores \gets \RelScore(r, R_{\text{act}})$\;
    $S \gets \emptyset$\;
    \ForEach{$e' \in E_{\text{act}}$}{
        $s \gets \MaxRelScore(e', r\_scores)$\;
        $S \gets S \cup \{(e', s)\}$\;
    }
    \Return{$S$} \tcp*[r]{Score connected entities}
}

\Fn{\AggregateGlobalScore($C$)}{
    \tcp{Calculate new scores and reassign for each $C\_e$ }
  \ForEach{$(C_e,\text{scores}) \in C$}{
    \ForEach{$(c, s) \in (C_e,\text{scores})$}{
      $s \gets \Sum([s'(c) \texttt{ for } (C', s') \texttt{ in } C \texttt{ if } c \in C'])$
    }
  }
  \Return{$C$}\;
}

\Fn{\RankTopK{$C, k_1$}}{
    $C^* \gets \emptyset$\;
    \ForEach{$(C_e,\text{scores}) \in C$}{
        $C_e^* \gets \TopKCand(C_e,\text{scores}, k_1)$\;
        $C^* \gets C^* \cup C_e^*$\;
    }
    \Return{$C^*$} \tcp*[r]{Rank candidates per unknown entity}
}

\Fn{\RetrieveComplete{$\mathcal{G}, T_{\text{comp}}, k_1, k_2$}}{
    $S \gets \emptyset$\;
    \ForEach{$(h,r,t)\in T_{\text{comp}}$}{
        \eIf{\ExistConn($h,t$)}{ 
            $T \gets \GetTripletsHT(\mathcal{G},h,t)$\;
            $scores \gets \emptyset$\;
            \ForEach{$t' \in T$}{
                $scores \gets scores \cup \RelScore(r, t'.\text{relation})$\;
            }
            $S \gets S \cup \TopK(T, scores, k_2)$ \tcp*[r]{Rank connected triplets}
        }{
            $S_1 \gets \RetrieveIncomplete(\mathcal{G}, [(unknown, r, t)], k_1)$\;
            $S_2 \gets \RetrieveIncomplete(\mathcal{G}, [(h, r, unknown)], k_1)$\;
            $S \gets S \cup (S_1 \cup S_2)$ \tcp*[r]{Handle missing connection}
        }
    }
    \Return{$S$}
}
\label{algo:subgraph_retrieval}
\end{algorithm*}

\section{Case Study}
\label{appendix:case_study}
We present the case study results of ClaimPKG on the FactKG dataset in Tables \ref{tab:correct_samples} and \ref{tab:incorrect_samples}. Each table includes the claim $c$, pseudo-subgraphs $P_s$, retrieved subgraphs $S_c$, final justification $j$, and verdict $v$. 
Table \ref{tab:correct_samples} showcases correctly predicted examples, demonstrating ClaimPKG’s ability to accurately capture claim structures and generate well-grounded justifications. Conversely, Table \ref{tab:incorrect_samples} highlights incorrectly predicted cases of two error types as detailed in Section \ref{error_analysis}. The first two examples illustrate Reasoning Errors, while the third represents a Retrieval Error. These insights serve as a foundation for future improvements, emphasizing key areas for future refinement.
\begin{table*}[h]
    \centering
    \renewcommand{\arraystretch}{1.2}
    \scriptsize
    \begin{adjustbox}{width=\textwidth}
    \begin{tabular}{p{2cm} p{5cm} p{5cm} p{2.5cm}}
        \toprule
        \toprule
        \multicolumn{1}{c}{\textbf{Claim $c$}} & 
        \multicolumn{1}{c}{\textbf{Pseudo-Subgraphs $P_c$}} & 
        \multicolumn{1}{c}{\textbf{Retrieved Subgraphs $S_c$}} & 
        \multicolumn{1}{c}{\textbf{\shortstack{Justification $j$; Verdict $v$}}} \\
        \midrule
        The 11th Mississippi Infantry Monument is not in Gettysburg, Pennsylvania. 
        & 
        \textbf{\textbf{P1:}} 11th Mississippi Infantry Monument || location || Gettysburg, Pennsylvania\newline
        \textbf{\textbf{P2:}} Gettysburg, Pennsylvania || \char`~location || 11th Mississippi Infantry Monument\newline
        11th Mississippi Infantry Monument || location || unknown\_0\newline
        \textbf{\textbf{P3:}} 11th Mississippi Infantry Monument || location || Gettysburg\newline
        11th Mississippi Infantry Monument || location || Pennsylvania \newline
        \textbf{\textbf{P4:}} 11th Mississippi Infantry Monument || is part of military conflict || Gettysburg\newline
        11th Mississippi Infantry Monument || location || Pennsylvania\newline
        11th Mississippi Infantry Monument || \char`~location || unknown\_0\newline
        \textbf{\textbf{P5:}} 11th Mississippi Infantry Monument || location || Gettysburg, Pennsylvania\newline
        &
        \textcolor{blue}{(11th Mississippi Infantry Monument, municipality, Gettysburg, Pennsylvania)}\newline
        (11th Mississippi Infantry Monument, website, 11th mississippi infantry monument)\newline
        (U.S. Route 140, location, Gettysburg)\newline
        (Pennsylvania Route 34, location, Gettysburg)\newline
        (U.S. Route 30 in Pennsylvania, location, Gettysburg)\newline
        \textcolor{blue}{(11th Mississippi Infantry Monument, state, Pennsylvania)}\newline
        (11th Mississippi Infantry Monument, region, Adams County, Pennsylvania)\newline
        (11th Mississippi Infantry Monument, country, United States)\newline
        (11th Mississippi Infantry Monument, category, Contributing property)\newline
        (Charles Saalmann, battles, Gettysburg)\newline
        (Lutheran Theological Seminary at Gettysburg, city, Gettysburg)\newline
        (Gettysburg Area High School, city, Gettysburg)\newline
        (11th Mississippi Infantry Monument, range, Seminary Ridge)
        &
        \textbf{Justification:} The triplets provided include '11th Mississippi Infantry Monument', 'municipality', 'Gettysburg, Pennsylvania' and '11th Mississippi Infantry Monument', 'state', 'Pennsylvania', indicating the monument's location as Gettysburg, Pennsylvania.\newline
        \textbf{Verdict:} \textbf{\textcolor{blue}{False}}
        \\
        \midrule
        It was released in Hardcover by J.V. Jones
        & 
        \textbf{\textbf{P1:}} Hardcover || \char`~media type || unknown\_0\newline
        unknown\_0 || author || J. V. Jones\newline
        \textbf{\textbf{P2:}} unknown\_0 || author || J. V. Jones\newline
        unknown\_0 || media type || Hardcover\newline
        unknown\_0 || \char`~genre || unknown\_1\newline
        \textbf{\textbf{P3:}} J. V. Jones || \char`~author || unknown\_0\newline
        unknown\_0 || media type || Hardcover\newline
        \textbf{\textbf{P4:}} Hardcover || \char`~media type || J. V. Jones\newline
        J. V. Jones || \char`~author || unknown\_0\newline
        unknown\_0 || previous work || unknown\_1\newline
        \textbf{\textbf{P5:}} Hardcover || \char`~media type || unknown\_0\newline
        unknown\_0 || author || J. V. Jones\newline
        &
        \textcolor{blue}{(Hardcover, \char`~media type, A Fortress of Grey Ice)}\newline
        (Hardcover, \char`~media type, A Clockwork Orange)\newline
        (Hardcover, \char`~media type, The Bell Curve)\newline
        \textcolor{blue}{(A Fortress of Grey Ice, author, J. V. Jones)}\newline
        (A Cavern of Black Ice, author, J. V. Jones)\newline
        (The Barbed Coil, author, J. V. Jones)\newline
        (Hardcover, \char`~media type, The Andromeda Strain)\newline
        (1963, \char`~date of birth, J. V. Jones)\newline
        (Liverpool, England, \char`~place of birth, J. V. Jones)\newline
        &
        \textbf{Justification:} The triplets show that 'A Fortress of Grey Ice' is a book by 'J. V. Jones' and it is of 'Hardcover' media type, which supports the claim.\newline
        \textbf{Verdict:} \textbf{\textcolor{blue}{True}}
        \\
        \midrule
        A person was born in the County of Tyrol and later died in Austria which is led by Doris Bures.
        & 
        \textbf{\textbf{P1:}} Austria || leader || Doris Bures\newline
        Austria || \char`~death place || unknown\_0\newline
        unknown\_0 || death place || County of Tyrol\newline
        \textbf{\textbf{P2:}} Austria || leader name || Doris Bures\newline
        Austria || \char`~death place || unknown\_0\newline
        unknown\_0 || death place || County of Tyrol\newline
        \textbf{\textbf{P3:}} County of Tyrol || \char`~birth place || unknown\_0\newline
        unknown\_0 || death place || Austria\newline
        unknown\_0 || successor || Doris Bures\newline
        \textbf{\textbf{P4:}} Doris Bures || \char`~leader || Austria\newline
        Austria || \char`~death place || unknown\_0\newline
        unknown\_0 || birth place || County of Tyrol\newline
        \textbf{\textbf{P5:}} Austria || leader || Doris Bures\newline
        Austria || \char`~death place || unknown\_0\newline
        unknown\_0 || birth place || County of Tyrol\newline
        &
        \textcolor{blue}{(Austria, leader, Doris Bures)}\newline
        (Austria, \char`~death place, Hans Hömberg)\newline
        (Austria, \char`~death place, Edgar Meyer (painter)\newline)\newline
        (Austria, \char`~death place, Rakhat Aliyev)\newline
        (Hans Hömberg, death place, County of Tyrol)\newline
        (Edgar Meyer (painter)\newline, death place, County of Tyrol)\newline
        (Friedrich Hasenöhrl, death place, County of Tyrol)\newline
        \textcolor{blue}{(County of Tyrol, \char`~birth place, Kurt Schuschnigg)}\newline
        \textcolor{blue}{(County of Tyrol, \char`~birth place, Alfons Gorbach)}\newline
        (County of Tyrol, \char`~birth place, Josef Hellensteiner)\newline
        \textcolor{blue}{(Kurt Schuschnigg, death place, Austria)}\newline
        \textcolor{blue}{(Alfons Gorbach, death place, Austria)}\newline
        (Josef Hellensteiner, death place, Austria)\newline
        (Barbara Prammer, successor, Doris Bures)\newline
        (Werner Faymann, successor, Doris Bures)\newline
        (Alois Stöger, \char`~successor, Doris Bures)\newline
        &
        \textbf{Justification:} The triplets show that Kurt Schuschnigg, Alfons Gorbach, and Josef Hellensteiner were born in the County of Tyrol and died in Austria. Additionally, Doris Bures is the leader of Austria.
        \textbf{Verdict:} \textbf{\textcolor{blue}{True}}
        \\
        \midrule
        The New Britain Transit is the operating organisation of Agra Airport which is located in India and has the location identifier VIAG.
        & 
        \textbf{\textbf{P1:}} Agra Airport || location || New Britain Transit\newline
        Agra Airport || location || India\newline
        Agra Airport || faa || VIAG\newline
        \textbf{\textbf{P2:}} Agra Airport || location country || India\newline
        Agra Airport || operator || New Britain Transit\newline
        Agra Airport || faa || VIAG\newline
        \textbf{\textbf{P3:}} Agra Airport || location || India\newline
        Agra Airport || operator || New Britain Transit\newline
        Agra Airport || faa || VIAG\newline
        \textbf{\textbf{P4:}} New Britain Transit || ~operator || Agra Airport\newline
        Agra Airport || location || India\newline
        Agra Airport || faa || VIAG\newline
        \textbf{\textbf{P5:}} Agra Airport || operator || New Britain Transit\newline
        Agra Airport || location || India\newline
        Agra Airport || faa || VIAG\newline
        &
        (Agra Airport, location, India)\newline
        (Agra Airport, location, Uttar Pradesh)\newline
        (Agra Airport, location, Agra)\newline
        \textcolor{blue}{(Connecticut Transit New Britain and Bristol, operator, New Britain Transit)}\newline
        \textcolor{blue}{(Agra Airport, icao location identifier, VIAG)}\newline
        \textcolor{blue}{(Agra Airport, operator, Indian Air Force)}\newline
        \textcolor{blue}{(Agra Airport, operator, Airports Authority of India)}\newline
        (Agra Airport, elevation, 167.94)\newline
        &
        \textbf{Justification:} The operator of Agra Airport is listed as 'Indian Air Force' and 'Airports Authority of India', not 'New Britain Transit', which operates 'Connecticut Transit New Britain and Bristol'\newline
        \textbf{Verdict:} \textbf{\textcolor{blue}{False}}
        \\
        \bottomrule
        \bottomrule
    \end{tabular}
    \end{adjustbox}
    \caption{Case study of ClaimPKG on the FactKG dataset with correctly predicted samples.}
    \label{tab:correct_samples}
\end{table*}
\begin{table*}[h]
    \centering
    \renewcommand{\arraystretch}{1.1} 
    \scriptsize
    \begin{adjustbox}{width=\textwidth}
    \begin{tabular}{p{1.8cm} p{4.6cm} p{5.2cm} p{2.5cm}}
        \toprule
        \toprule
        \multicolumn{1}{c}{\textbf{Claim $c$}} & 
        \multicolumn{1}{c}{\textbf{Pseudo-Subgraphs $P_c$}} & 
        \multicolumn{1}{c}{\textbf{Retrieved Subgraphs $S_c$}} & 
        \multicolumn{1}{c}{\textbf{\shortstack{Justification $j$; Verdict $v$}}} \\
        \midrule
        When did Indian Air Force operate Agra Airport?
        & 
        \textbf{P1:} Agra Airport || operator || Indian Air Force\newline
        \textbf{P2:} Indian Air Force || \char`~operator || Agra Airport\newline
        Agra Airport || \char`~runway designation || unknown\_0\newline
        \textbf{P3:} Agra Airport || \char`~runway designation || Indian Air Force\newline
        Agra Airport || \char`~location || unknown\_0\newline
        unknown\_0 || operator || Indian Air Force\newline
        \textbf{P4:} Agra Airport || operator || Indian Air Force\newline
        \textbf{P5:} Agra Airport || operator || Indian Air Force\newline
        &
        (Agra Airport, operator, Indian Air Force)\newline
        (Agra Airport, runway length, 2743.0)\newline
        (Agra Airport, runway length, 1818.0)\newline
        (Agra Airport, icao location identifier, VIAG)\newline
        (Agra Airport, location, India)\newline
        (Agra Airport, location, Uttar Pradesh)\newline
        (Agra Airport, location, Agra)\newline
        (Bhuj Airport, operator, Indian Air Force)\newline
        (Bhuj Rudra Mata Air Force Base, operator, Indian Air Force)\newline
        (Car Nicobar Air Force Base, operator, Indian Air Force)\newline
        &
        \textbf{Justification:} The triplets do not provide information about the specific time period when the Indian Air Force operated Agra Airport. However, they do confirm that the Indian Air Force is the operator of Agra Airport, implying that the operation is either current or has occurred at some point, but the exact timing is not specified.\newline
        \textbf{Verdict: \textcolor{red}{False}}
        \\
        \midrule
        I imagined that Lady Anne Monson not lived in the Cape Verde whose capital was Winchester.
        &
        \textbf{P1:} Cape Verde || capital || Winchester\newline
        Cape Verde || \char`~residence || Lady Anne Monson\newline
        \textbf{P2:} Lady Anne Monson || residence || Cape Verde\newline
        Cape Verde || capital || Winchester\newline
        \textbf{P3:} Anne Monzon || residence || Cape Verde\newline
        Cape Verde || capital || Winchester\newline
        \textbf{P4:} Cape Verde || capital || Winchester\newline
        Cape Verde || \char`~residence || Lady Anne Monson\newline
        \textbf{P5:} Cape Verde || capital || Winchester\newline
        Cape Verde || \char`~death place || Lady Anne Monson\newline
        &
        (Cape Verde, capital, Praia)\newline
        (Cape Verde, currency, Cape Verdean escudo)\newline
        (Cape Verde, \char`~headquarter, Movement for Democracy (Cape Verde))\newline
        (Wessex, capital, Winchester)\newline
        (Kingdom of England, capital, Winchester)\newline
        (River Itchen, Hampshire, city, Winchester)\newline
        (Cape Verde, \char`~birth place, Aristides Pereira)\newline
        (Cape Verde, \char`~birth place, Pedro Pires)\newline
        (Cape Verde, \char`~birth place, José Maria Neves)\newline
        (India, \char`~residence, Lady Anne Monson)\newline
        (Kingdom of England, \char`~birth place, Lady Anne Monson)\newline
        (Darlington, \char`~birth place, Lady Anne Monson)\newline
        (Anne Monzon, \char`~cinematography, "Wont Last a Day Without You")\newline
        (Anne Monzon, \char`~cinematography, Pop Class)\newline
        (Anne Monzon, \char`~cinematography, Namets!)\newline
        (Cape Verde, \char`~death place, Corsino Fortes)\newline
        (Cape Verde, \char`~death place, Cesária Évora)\newline
        (Cape Verde, \char`~death place, Baltasar Lopes da Silva)\newline
        (1776-02-18, \char`~death date, Lady Anne Monson)\newline
        (1776, \char`~death year, Lady Anne Monson)\newline
        &
        \textbf{Justification:} The capital of Cape Verde is Praia, not Winchester. Winchester is the capital of Wessex and the Kingdom of England. Lady Anne Monson's residence is listed as India, and her birthplace is listed as Darlington and the Kingdom of England, with no mention of Cape Verde as her residence.\newline
        \textbf{Verdict: \textcolor{red}{True}} 
        \\
        \midrule
        The operating organization for Agra Airport is the Indian Air Force who deploy the HAL Light Combat Helicopter and the Boeing C17 Globemaster III transport aircraft.
        & 
        \textbf{P1:} Agra Airport || operator || Indian Air Force\newline
        Agra Airport || \char`~location || Boeing C-17 Globemaster III\newline
        Agra Airport || \char`~location || HAL Light Combat Helicopter\newline
        \textbf{P2:} Boeing C-17 Globemaster III || \char`~aircraft || Agra Airport\newline
        Agra Airport || operator || Indian Air Force\newline
        HAL Light Combat Helicopter || \char`~aircraft || Agra Airport\newline
        \textbf{P3:} Agra Airport || operating organization || Indian Air Force\newline
        Agra Airport || \char`~location || Boeing C-17 Globemaster III\newline
        Agra Airport || \char`~location || HAL Light Combat Helicopter\newline
        \textbf{P4:} Agra Airport || operator || Indian Air Force\newline
        Agra Airport || runway length || Boeing C-17 Globemaster III\newline
        Agra Airport || \char`~location || HAL Light Combat Helicopter\newline
        \textbf{P5:} Agra Airport || operator || Indian Air Force\newline
        Agra Airport || \char`~city || HAL Light Combat Helicopter\newline
        Agra Airport || \char`~city || Boeing C-17 Globemaster III\newline
        &
        (Agra Airport, operator, Indian Air Force)\newline
        (Agra Airport, location, India)\newline
        (Agra Airport, location, Uttar Pradesh)\newline
        (Agra Airport, location, Agra)\newline
        (2012 in spaceflight, site, Boeing C-17 Globemaster III)\newline
        (2013 in spaceflight, site, Boeing C-17 Globemaster III)\newline
        (2009 in spaceflight (July–December)\newline, site, Boeing C-17 Globemaster III)\newline
        (India, ~origin, HAL Light Combat Helicopter)\newline
        (Developmental trials, ~status, HAL Light Combat Helicopter)\newline
        (Hindustan Aeronautics, ~manufacturer, HAL Light Combat Helicopter)\newline
        (Boeing C-17 Globemaster III, ~aircraft transport, United States Air Force)\newline
        (Boeing C-17 Globemaster III, ~aircraft transport, Royal Air Force)\newline
        (Boeing C-17 Globemaster III, ~aircraft transport, Royal Australian Air Force)\newline
        (2743.0, ~runway length, Agra Airport)\newline
        (1818.0, ~runway length, Agra Airport)\newline
        (HAL Light Combat Helicopter, ~aircraft helicopter, Indian Air Force)\newline
        (HAL Light Combat Helicopter, ~aircraft attack, Army Aviation Corps (India))\newline
        (HAL Light Combat Helicopter, type, Attack helicopter)\newline
        (Agra Airport, icao location identifier, VIAG)\newline
        (Airlift, ~type, Boeing C-17 Globemaster III)\newline
        (United States, ~origin, Boeing C-17 Globemaster III)\newline
        (In service, ~status, Boeing C-17 Globemaster III)\newline
        &
        \textbf{Justification: }The triplets confirm Agra Airport is operated by the Indian Air Force, the Indian Air Force deploys the HAL Light Combat Helicopter, and the Boeing C-17 Globemaster III is indeed a transport aircraft, but there is no direct information linking the Indian Air Force at Agra Airport to the deployment of the Boeing C-17 Globemaster III. However, given the information provided, we can infer the involvement of the Indian Air Force with the mentioned aircraft types, albeit indirectly for the Boeing C-17 Globemaster III.\newline
        \textbf{Verdict: \textcolor{red}{False}} 
        \\
        \bottomrule
        \bottomrule
    \end{tabular}
    \end{adjustbox}
    \caption{Case study of ClaimPKG on the FactKG dataset with incorrectly predicted samples.}
    \label{tab:incorrect_samples}
\end{table*}

\section{Prompt Templates}
For better reproducibility, we present all prompt templates in the appendix. Below is a quick reference list outlining the prompt templates and their usages:
\begin{itemize}
    \item Figure \ref{fig:prompt_general_reasoning}: Prompt the General LLM to reason on the input claim and retrieved subgraphs to produce justification and final verdict. 
    \item Figure \ref{fig:prompt_fewshot_pseudo_subgraph}: Few-shot prompts the General LLM to generate a Pseudo Subgraph with provided examples.
    \item Figure \ref{fig:prompt_in_out_entities}: Annotate the inside and outside entities of the input claim for the training dataset.
\end{itemize}


\begin{figure*}[!h]
    \centering
    \begin{tcolorbox}[
        colback=gray!10,
        boxrule=0.4pt,
        arc=2pt,
        left=4pt,     
        right=4pt,    
        top=4pt,       
        bottom=4pt,     
        fontupper={\small}
    ]
    \textbf{GENERAL REASONING} \\ \\
    \textbf{Task:} \\
    Verify whether the fact in the given sentence is true or false based on the provided graph triplets. Use only the information in the triplets for verification. \\
    \\
    - The triplets provided represent all relevant knowledge that can be retrieved. \\
    - If the fact is a negation and the triplets do not include the fact, consider the fact as true. \\
    - Ignore questions and verify only the factual assertion within them. For example, in the question ``When was Daniel Martínez (politician) a leader of Montevideo?'', focusing on verifying the assertion ``Daniel Martínez (politician) a leader of Montevideo''. \\
    - Interpret the ``$\sim$'' symbol in triplets as indicating a reverse relationship. For example: ``A $\sim$ south of B'' means ``B is north of A''. \\
    \\
    \textbf{Response Format:} \\
    Provide your response in the following JSON format without any additional explanations: \\
    \{ \\
    \text{\quad "rationale": "A concise explanation for your decision",} \\
    \text{\quad "verdict": "true/false as the JSON value"} \\
    \} \\
    \\
    \textbf{Triplets:} \\
    \{\{triplets\}\} \\
    \\
    \textbf{Claim:} \\
    \{\{claim\}\}
    \end{tcolorbox}
    \caption{Prompt template for the general LLM to perform reasoning}
    \label{fig:prompt_general_reasoning}
\end{figure*}


\begin{figure*}[!h]
    \centering
    \begin{tcolorbox}[
        colback=gray!10,
        boxrule=0.4pt,
        arc=2pt,
        left=4pt,     
        right=4pt,    
        top=4pt,       
        bottom=4pt,     
        fontupper={\small}
    ]
    \textbf{FEWSHOT PSEUDO SUBGRAPH GENERATION} 
    \\ \\
    \textbf{Task:} Generate a reference graph to verify the following claim. Only return the subgraphs following the format of provided examples and do NOT include other unnecessary information.
    \\ \\
    \textbf{Here are some examples:}
    \\ \\
    \textbf{Claim:} Akeem Priestley played for club RoPS and currently plays for the Orange County Blues FC, which is managed by Oliver Wyss.\\
    \textbf{Subgraphs:}\\
    <e>Orange County Blues FC</e> || manager || <e>Oliver Wyss</e>\\
    <e>Orange County Blues FC</e> || ~clubs || <e>Akeem Priestley</e>\\
    <e>Akeem Priestley</e> || team || <e>RoPS</e>
    \\ \\
    \textbf{Claim:} He is a Rhythm and Blues singer from Errata, Mississippi!\\
    \textbf{Subgraphs:}\\
    <e>Rhythm and blues</e> || ~genre || unknown\_0\\
    unknown\_0 || birth place || <e>Errata, Mississippi</e>\\
    unknown\_0 || background || unknown\_1
    \\ \\
    \textbf{Claim:} Arròs negre is a traditional dish from Spain, and from the Catalonia region, which is led by the Maria Norrfalk.\\
    \textbf{Subgraphs:}\\
    <e>Arròs negre</e> || country || <e>Spain</e>\\
    <e>Arròs negre</e> || region || <e>Catalonia</e>\\
    <e>Catalonia</e> || leader name || <e>Maria Norrfalk</e>
    \\ \\
    \textbf{Claim:} Well, Jason Sherlock did not have a nickname!\\
    \textbf{Subgraphs:}\\
    <e>Jason Sherlock</e> || nickname || unknown\_0
    \\ \\
    \textbf{Claim:} Garlic is the main ingredient of Ajoblanco, which is from Andalusia.\\
    \textbf{Subgraphs:}\\
    <e>Ajoblanco</e> || region || <e>Andalusia</e>\\
    <e>Ajoblanco</e> || ingredient || <e>Garlic</e>
    \\ \\
    ..... More examples .....
    \\ \\
    \textbf{Claim:} \{\{claim\}\} \\
    \textbf{Subgraphs:} \\
    \end{tcolorbox}
    \caption{Prompt template for the general LLM to generate pseudo subgraphs}
    \label{fig:prompt_fewshot_pseudo_subgraph}
\end{figure*}


\begin{figure*}[!h]
    \centering
    \begin{tcolorbox}[
        colback=gray!10,
        boxrule=0.4pt,
        arc=2pt,
        left=4pt,     
        right=4pt,    
        top=4pt,       
        bottom=4pt,     
        fontupper={\small}
    ]
    \textbf{ANNOTATE IN AND OUT ENTITIES}
    \\ \\
    \textbf{Task:} Specify if the following entities are mentioned in the claim or not. \\
    Respond correctly in the following JSON format and do not output anything else: \\
    \{ \\
        \text{\quad "in\_entities": [list of entities that are in the claim],} \\
        \text{\quad "out\_entities": [list of entities that are not in the claim]} \\
    \} \\
    Do not change the entity names from the list of provided entities. \\
    
    \textbf{Claim:} \{\{claim\}\} \\
    \textbf{Entities:} \{\{entities\}\}
    \end{tcolorbox}
    \caption{Prompt template to annotate inside and outside entity of the claim.}
    \label{fig:prompt_in_out_entities}
\end{figure*}

\end{document}